\documentclass[lettersize,journal]{IEEEtran}
\usepackage{amsmath,amsfonts}
\usepackage{algorithmic}
\usepackage{algorithm}
\usepackage{array}
\usepackage[caption=false,font=footnotesize,labelfont=sf,textfont=sf]{subfig}
\usepackage{textcomp}
\usepackage{stfloats}
\usepackage{url}
\usepackage{verbatim}
\usepackage{graphicx}
\usepackage{cite}
\usepackage{booktabs}
\usepackage{amsthm}
\usepackage{multicol}
\usepackage{multirow}
\usepackage{diagbox}
\newtheorem{proposition}{Proposition}
\newtheorem{lemma}{Lemma}
\begin{document}
\renewcommand{\algorithmicrequire}{\textbf{Input:}}  
\renewcommand{\algorithmicensure}{\textbf{Output:}}  

\title{IPF-RDA: An Information-Preserving Framework for Robust Data Augmentation}

\author{Suorong~Yang*, 
	Hongchao~Yang,
	Suhan~Guo,
	Furao~Shen*\thanks{* Corresponding author.},
        and Jian~Zhao
 
	\thanks{Suorong~Yang and Hongchao~Yang is with State Key Laboratory for Novel Software Technology, Department of Computer Science and Technology, Nanjing University, Nanjing 210023, China (e-mail: sryang@smail.nju.edu.cn; yanghc@smail.nju.edu.cn)}
	\thanks{Suhan~Guo and Furao~Shen are with State Key Laboratory for Novel Software Technology, Nanjing University, China, School of Artificial Intelligence, Nanjing University, Nanjing 210023, China (e-mail: shguo@smail.nju.edu.cn; frshen@nju.edu.cn) }
	\thanks{Jian Zhao is with the School of Electronic Science and Engineering, Nanjing University, Nanjing 210023, China (e-mail: jianzhao@nju.edu.cn)}
}
\maketitle

\begin{abstract}
Data augmentation is widely utilized as an effective technique to enhance the generalization performance of deep models. However, data augmentation may inevitably introduce distribution shifts and noises, which significantly constrain the potential and deteriorate the performance of deep networks. To this end, we propose a novel information-preserving framework, namely IPF-RDA, to enhance the robustness of data augmentations in this paper. IPF-RDA combines the proposal of (i) a new class-discriminative information estimation algorithm that identifies the points most vulnerable to data augmentation operations and corresponding importance scores; And (ii) a new information-preserving scheme that preserves the critical information in the augmented samples and ensures the diversity of augmented data adaptively. We divide data augmentation methods into three categories according to the operation types and integrate these approaches into our framework accordingly. After being integrated into our framework, the robustness of data augmentation methods can be enhanced and their full potential can be unleashed. Extensive experiments demonstrate that although being simple, IPF-RDA consistently improves the performance of numerous commonly used state-of-the-art data augmentation methods with popular deep models on a variety of datasets, including CIFAR-10, CIFAR-100, Tiny-ImageNet, CUHK03, Market1501, Oxford Flower, and MNIST, where its performance and scalability are stressed. The implementation is available at \url{https://github.com/Jackbrocp/IPF-RDA}.

\end{abstract}

\begin{IEEEkeywords}
Robust data augmentation, deep learning, model generalization, interpretability.
\end{IEEEkeywords}

\section{Introduction}
Data augmentation (DA) has emerged as an effective technique for enhancing model optimization by providing diverse training data, which has been widely used in various tasks~\cite{survey0,survey1,survey2}.
Recently, many DA methods~\cite{cutout,cutmix,autoaugment, randaugment, trivialaugment,keepaugment,advmask,yang2025dynamic} have been proposed, achieving noteworthy results.
Despite the effectiveness, recent research has underscored some challenges with DA: 1) Existing DA techniques face a dilemma between diversifying the data and maintaining its consistency. 
While DA methods can bolster model performance, they might simultaneously introduce distribution shifts, compromising performance on non-augmented data during inference~\cite{selectaugment,keepaugment,trivialaugment,yang2024clip}.
Particularly concerning is the latent risk that DA may distort critical information for classification, resulting in mislabeled or ambiguously labeled augmented data.
2) DA methods primarily focus on globally transforming images using a unified strategy, with reduced attention given to the local discriminative context in individual images.
This may result in a bottleneck for performance improvement~\cite{advmask,ma2022sage}.
Consequently, these observations highlight the limitations of existing DA techniques, suggesting that inattentively designed augmentations could subtly but adversely affect the generalization capability of neural models. 
To avoid this, KeepAugment~\cite{keepaugment} has recently been proposed, utilizing the saliency map to assess the importance of pixels in images.
At each training iteration, it performs data augmentation by keeping these salient points untouched.
Although KeepAugment has achieved improvements over Cutout~\cite{cutout}, CutMix~\cite{cutmix}, AutoAugment~\cite{autoaugment}, and RandAugment~\cite{randaugment}, it is based on a basic assumption that the saliency map accurately identifies the points most sensitive to the classification decision; thus, these points are also the most vulnerable ones to the influence introduced by the DA operations.
However, some publications have empirically shown that saliency methods do not provide insightful explanations to the classification decisions and cannot be used as the basis for classification~\cite{saliency, saliency_reliabiliey, SalientDN, sanity}, rendering them non-class-discriminative.
Meanwhile, the online importance scoring mechanism also introduces considerable computational overhead to the training process.
To address this important but rarely studied challenge, an imminent problem arises: \textit{How can we preserve the class-discriminative information without sacrificing the diversity of augmented data brought about by DA, hence enhancing its robustness?}
By enhancing the methods' robustness, we can further improve the efficacy of DA methods and unleash their full potential.


In this work, we propose a novel Information-Preserving Framework for Robust Data Augmentation (IPF-RDA).
This framework is motivated by the intriguing observation made by recent work showing that data augmentation may introduce distribution shift and consequently hurt the performance on unaugmented data during inference~\cite{keepaugment, trivialaugment}.
The idea of our method is straightforward yet effective: IPF-RDA identifies the most class-discriminative information in an offline manner and preserves this information while maintaining the diversity of augmented data during training.
To achieve this, first, we theoretically prove that the vanilla saliency map is not sensitive to the classification results.
Based on the theoretical proof, we introduce our offline class-discriminative information estimation algorithm (CDIEA), which identifies the points at which minor perturbations lead to misclassification.
These points represent the most vulnerable features to data augmentation operations.
Leveraging these critical points and their associated importance scores, IPF-RDA adaptively preserves a randomly selected rectangular region containing the class-discriminative information in each epoch during online training.
The use of rectangular preservation ensures that the most critical features remain intact while maintaining the structural semantic characteristics of the object and its local context~\cite{gridmask,advmask,keepaugment,cutmix,cutout}, combining both practical efficiency and effectiveness.
Furthermore, by departing the CDIEA from the online training process, our framework imposes minimal additional training overhead during the online augmentation phase.
Notably, our framework is highly scalable as it can significantly improve a number of state-of-the-art (SOTA) and widely adopted DA methods, including but not limited to Cutout~\cite{cutout}, AdvMask~\cite{advmask}, GridMask~\cite{gridmask}, Hide-and-Seek (HaS)~\cite{has}, CutMix~\cite{cutmix}, RandomErasing~\cite{random_erasing}, AutoAugment~\cite{autoaugment}, Fast-AutoAugment~\cite{fast_autoaugment}, RandAugment~\cite{randaugment}, TrivialAugment~\cite{trivialaugment}, etc.
Consequently, the main benefit of our framework lies in its ability to enhance robustness, improve the performance of data augmentation methods, and unleash their full potential, thereby facilitating optimized model training.  
Although the idea is simple, the proposed IPF-RDA is significantly effective and easy to implement.
Extensive empirical evaluations, including supervised and semi-supervised image classification on CIFAR-10/100~\cite{cifar}, and Tiny-ImageNet~\cite{tiny}, person re-identification (ReID) on CUHK03~\cite{cuhk03} and Market1501~\cite{market1501}, and a series of qualitative and quantitative analyses are conducted.
Experimental results demonstrate that IPF-RDA consistently improves the generalization performance of representative and widely used SOTA DA methods.

In summary, our contributions are:
\begin{itemize}
    \item We introduce an information-preserving framework designed to integrate the most representative and widely used data augmentation methods, thereby further improving their performance by reducing the side effects caused by the augmentation operations, which are overlooked in prior works. 
    \item Our study begins with a theoretical clarification highlighting that the vanilla saliency map is not sensitive to classification determinants. Next, we propose an offline algorithm for estimating class-discriminative information, which is designed to enhance data augmentation robustness.
    \item We develop a taxonomy that categorizes methods based on their operational types and subsequently integrates DA methods into our information-preserving framework according to their categories.
    \item Experimental results on various standard benchmark datasets demonstrate the efficacy of our framework. IPF-RDA can be employed to effectively enhance the robustness of SOTA DA methods, thus significantly improving their performance on various neural architectures.
\end{itemize}
\begin{figure}[]
    \centering
    \includegraphics[width=0.8\columnwidth]{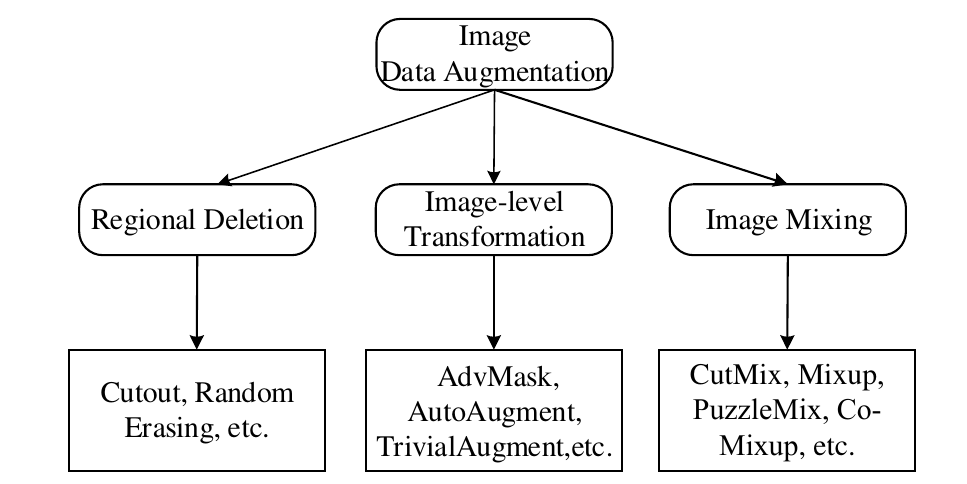}
    \caption{A new taxonomy of image data augmentation approaches.}
    \label{taxnomy}
\end{figure}
 \section{Related Work}
Over recent years, many image data augmentation methods have been proposed based on different perspectives of image manipulations.
As illustrated in Fig.~\ref{taxnomy}, we introduce a new taxonomy that categorizes data augmentation approaches into three types: regional deletion, image-level transformation, and image mixing. 

\textbf{Regional deletion} techniques, such as Cutout~\cite{cutout} and Random Erasing~\cite{random_erasing}, have been commonly employed.
 Cutout~\cite{cutout} is a simple regularization technique that randomly masks out rectangular regions of input images.
 Random Erasing ~\cite{random_erasing} aims to enhance the model's robustness to occlusion by randomly selecting a rectangular region in an image and replacing its pixels with random values.
 Despite the benefits of augmenting images with diverse positions and sizes of deleted regions, there exists a significant risk of eliminating critical areas during the augmentation process~\cite{keepaugment,advmask}.
 
 To address this concern, our proposed framework integrates these methods but prevents them from removing highly class-discriminative regions, thereby enhancing the robustness of regional deletion-based DA approaches.

\textbf{Image-level transformation} includes image-level deletion-based methods such as Hide-and-Seek (HaS), GridMask~\cite{gridmask}, AdvMask~\cite{advmask}, and many automatic augmentation methods.
HaS involves randomly hiding multiple patches within training samples.
GridMask~\cite{gridmask} utilizes structured dropping regions, implicitly emphasizing the importance of balancing information reservation and deletion.
AdvMask~\cite{advmask} first utilizes sparse adversarial attack points as key points and randomly masks some of them to force models to learn discriminative information in the images while some critical information is missing.
Unlike AdvMask, our approach applies adaptive importance scores to evaluate the significance of critical points, indicating different levels of importance.
Nevertheless, to foster data augmentation diversity, the masked regions of these methods are randomly generated.
Although these methods can enhance the robustness of models to occlusions, such random deletion may inadvertently remove discriminative information crucial for classification or even delete objects of interest.

Automatic augmentation includes AutoAugment~\cite{autoaugment}, Fast-AutoAugment~\cite{fast_autoaugment}, RandAugment~\cite{randaugment}, and TrivialAugment~\cite{trivialaugment}, etc.
These methods primarily leverage reinforcement learning to explore optimal augmentation strategies and corresponding parameters for each dataset in an offline manner.
During the augmentation process, these methods randomly sample augmentation policies and strength settings in augmentation spaces.
For instance, both AutoAugment~\cite{autoaugment} and Fast-AutoAugment~\cite{fast_autoaugment} define the problem of finding the best augmentation policies as a discrete search problem and automatically search for the optimal combination of DA policies via reinforcement learning.
RandAugment~\cite{randaugment} adopts a significantly reduced search space, allowing it to be trained on the target task without a separate proxy task.
To maintain the diversity of augmented data~\cite{yang2024investigating}, these methods simultaneously apply different augmentation policies.
However, the simultaneous application of multiple operations on a single image can potentially introduce distribution discrepancies~\cite{trivialaugment,keepaugment,yang2024entaugment}.
In contrast, TrivialAugment~\cite{trivialaugment} adopts an empirical approach, applying a single augmentation to each image.
By employing a single augmentation policy, the rationale of TrivialAugment is to implicitly mitigate the impact of augmentation operations on images, effectively preserving the original image information.

We integrate image-level transformation-based DA methods into our proposed framework by explicitly recovering the critical information in images regardless of the extent to which image-level transformation methods alter images.
Therefore, our framework is capable of enhancing the robustness of these DA methods while maintaining the diversity inherent in the original augmentation processes, thereby fully harnessing their potential.

\textbf{Image mixing}, such as Mixup~\cite{mixup} and CutMix~\cite{cutmix}, has gained attention in the field of data augmentation.
PuzzleMix~\cite{puzzlemix}, a mixup method, explicitly utilizes the saliency information and the underlying statistics of the natural examples to enhance generalization. 
Meanwhile, Co-Mixup~\cite{co-mixup} introduces a new perspective on batch mixup by formulating an optimization objective that maximizes the data saliency of each individual mixup sample while encouraging supermodular diversity across the batch.
These methods combine two images or sub-regions. Meanwhile, labels of the augmented data are fused based on the area proportions of the replaced regions. 
However, a challenge arises when critical regions essential for classification are affected or covered during the mixing process.
This leads to ambiguous labels for the augmented data.
For example, consider two images $x_a$ and $x_b$, along with the corresponding labels $y_a$ and $y_b$.
CutMix randomly selects a region in $x_a$ and replaces it with the corresponding region from $x_b$:
\begin{equation}\label{eq:image-mixing}
    \begin{aligned}
& \tilde{x}=\mathbf{M} \odot x_a+(\mathbf{1}-\mathbf{M}) \odot x_b \\
& \tilde{y}=\omega y_a+(1-\omega) y_b
\end{aligned}
\end{equation}
where $\mathbf{M} \in\{0,1\}^{W \times H}$ denotes a binary mask indicating where to drop out or fill in from two images (i.e., filling with 0 within the selected region, otherwise with 1), $\mathbf{1}$ is a binary mask filled with ones.
Suppose that the selected region is of size $w_M\times h_M$, $\omega=\frac{w_M h_M}{WH}$ is determined as the proportion of the area of $x_b$ in the composite image $\tilde{x}$.
However, this kind of combination may inevitably introduce noisy samples when the area of the selected region is large, but there is no object from label $y_b$ in that region. 
In this way, the augmented label is mainly $y_b$ while the augmented image only contains objects of $y_a$.

To address this issue, our framework integrates CutMix by first replacing a less important region in $x_a$ with a corresponding region from $x_b$ to preserve the information from $y_a$.
Meanwhile, we improve the calculation of $\omega$ in our framework.
Instead of relying on the area proportions, the weights of labels are calculated based on the importance scores of the selected regions.

\textbf{Information preserving} has been implicitly applied in more recent data augmentation methods~\cite{trivialaugment,selectaugment,advmask}, while KeepAugment~\cite{keepaugment} first explicitly preserves salient regions in the augmented data.
Specifically, KeepAugment focuses on enhancing the fidelity of augmented images by leveraging the vanilla saliency map to identify informative regions in the original images.
These informative regions are preserved during the augmentation process.
However, it has been shown empirically that the vanilla saliency map is not sensitive to the classification decision~\cite{saliency, saliency_reliabiliey, SalientDN, sanity}.
Meanwhile, the calculation of the saliency map in each iteration increases the training costs of the augmentation process.
Instead, our framework utilizes an offline algorithm, CDIEA, to identify the most class-discriminative regions.
Compared with KeepAugment, the proposed framework can improve upon a larger number of DA methods, including but not limited to Cutout, Random Erasing, HaS, GridMask, AdvMask, AutoAugment, RandAugment, TrivialAugment, etc., thus demonstrating higher generality.

Another line of related work is based on deep generative models~\cite{dgm,dgm2}.
By training a generative adversarial network, one could utilize samples from the generator as the augmented data.
Unfortunately, their adoption remains limited within the data augmentation research community due to significant computational requirements and the need for substantial training data.
Specifically, the complexity associated with training generative networks and inferring augmented samples from them presents nontrivial challenges, limiting the widespread use of these methods~\cite{isda}.
\section{Proposed Method}
\begin{figure}[]
    \centering
    \includegraphics[width=0.95\columnwidth]{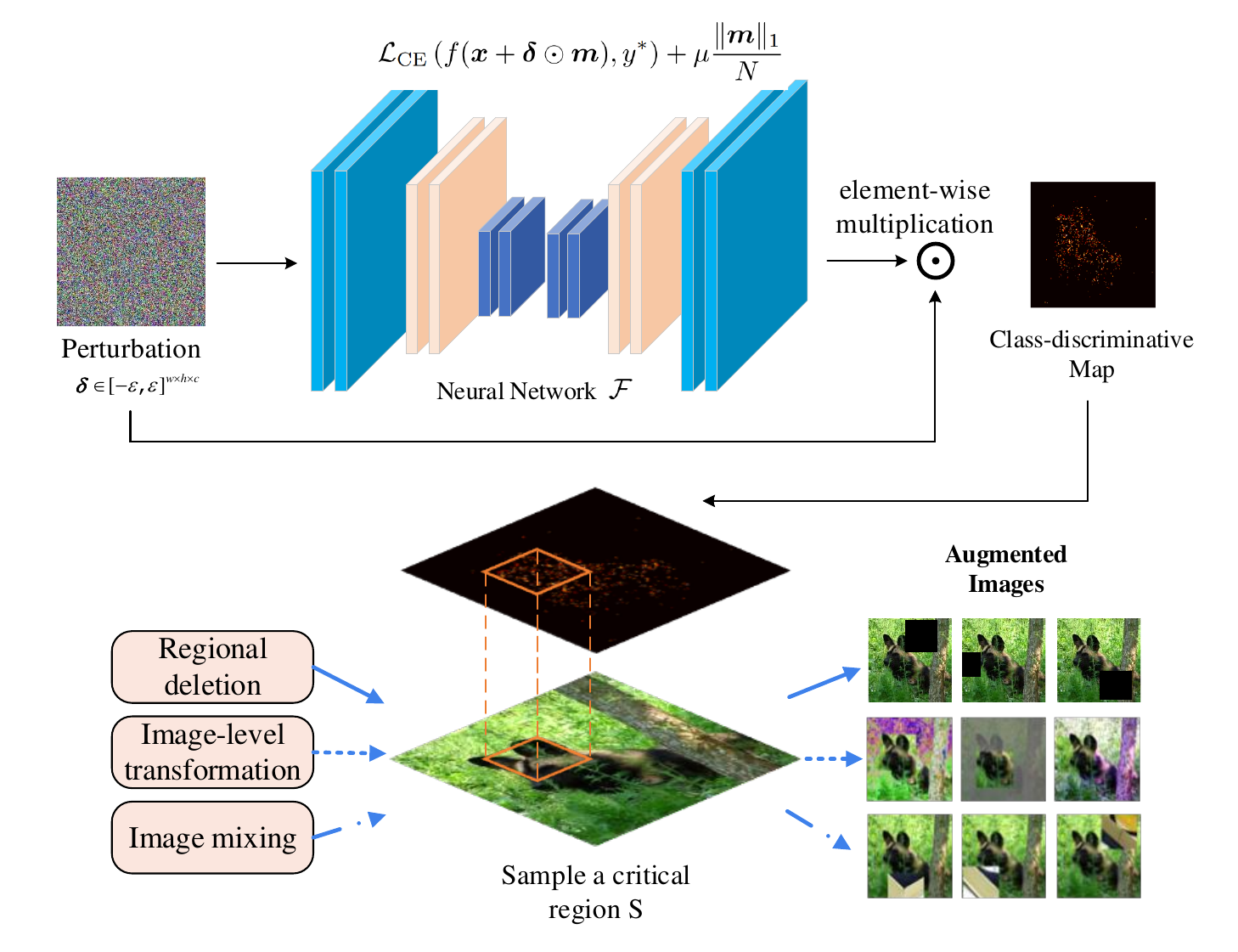}
    \caption{Overview of the proposed framework. }
    \label{framework}
\end{figure}
As illustrated in Fig.~\ref{framework}, the proposed method consists of two components: the Class-Discriminative Information Estimation Algorithm (CDIEA) in the upper portion and the Information-Preserving Framework (IPF) in the lower portion of the diagram.
The former, CDIEA, operates offline and is designated to identify the most class-discriminative information (i.e., the positions and importance scores of critical pixels).
Concurrently, IPF is responsible for preserving this critical information, which ensures robust augmentations throughout online training.
In this section, we first provide a theoretical analysis in Section~\ref{sec:analysis} to show that saliency methods are unsuitable for identifying class-discriminative information.
Based on the theoretical analysis, we describe our proposed end-to-end CDIEA in Section~\ref{sec:SAA}.
Subsequently, in Section~\ref{sec:IPF}, we describe how classification-sensitive information can be utilized to enhance the robustness of different types of DA techniques.
\subsection{Theoretical Analysis}\label{sec:analysis}
Although KeepAugment shares a similar concept of preserving critical regions for improved performance, empirical studies from~\cite{SM1,SM2,SM3} have demonstrated that saliency maps exhibit a lack of sensitivity to classification decisions.
In this section, we offer theoretical substantiation to corroborate this observation by establishing that conventional gradient-based saliency maps do not possess class-discriminative properties.

The saliency map involves computing the gradient of the loss function for the class of interest w.r.t. the input pixels $x$, denoted as $g_x = \frac{\partial \mathcal{L}_c(x)}{\partial x}$.
To assess the sensitivity of the saliency map to classification decisions, we consider a successful adversarial attack on the model $f$ by applying a perturbation $\boldsymbol{\delta}$ on the input image $\boldsymbol{x}$ of class $y$, that is, $f(\boldsymbol{x}+\boldsymbol{\delta}) \neq y$.
Therefore, this perturbation $\boldsymbol{\delta}$ has already modified the classification decision.
Subsequently, we proceed to examine the saliency map of an adversarial sample $\boldsymbol{x}+\boldsymbol{\delta}$ using a ReLU network $f$.
Before going into the main results, we need the following proposition in~\cite{whySAA} for our proof.
\begin{proposition}\label{prop:whySAA}
\cite{whySAA} Let $\lambda_i$ and $v_i$ be the $i$-th largest eigenvalue of the Hessian matrix and its corresponding eigenvector, respectively.
$\gamma_i \in \mathbb{R}$ is the projection length of $g_x$ on $v_i$, and $\beta$ is the update step used to obtain the adversarial sample. We have, 
\begin{equation}
   \boldsymbol{\delta} =\sum_i \frac{\left(1+\beta \lambda_i\right)^m-1}{\lambda_i} \gamma_i v_i+\mathcal{R}_2(\beta)
\end{equation}
\end{proposition}
\begin{lemma}\label{lemma1}
 Let us assume the adversarial attack is accomplished in $m$ steps. 
We establish the relationship between the saliency maps of sample $\boldsymbol{x}$ and the adversarial sample $\boldsymbol{x}+\boldsymbol{\delta}$ in $f$ as follows:
 \begin{equation}
    g_{\boldsymbol{x}+\boldsymbol{\delta}}  = g_{\boldsymbol{x}} + H_x\sum_i \frac{\left(1+\beta \lambda_i\right)^m-1}{\lambda_i} \gamma_i v_i +H_x\mathcal{R}_2(\beta) 
 \end{equation}
 where $H_x$ is the Hessian matrix defined as $H_x \stackrel{\text{def}}{=} \frac{\partial^2}{\partial x \partial x^T} \mathcal{L}(f(x), y)$ and $\mathcal{R}_2(\beta)$ represents the terms of $\beta$ that are at least of second order.
\end{lemma}

Lemma~\eqref{lemma1} is proven in Appendix~\ref{appendix:lemma}.
Based on Lemma~\ref{lemma1}, we can see that the gradient of the adversarial sample can be approximately derived based on the model parameters regardless of specific $\boldsymbol{\delta}$.
Furthermore, the difference in the saliency map of the adversarial sample and the initial sample can be approximated as $\Delta_g=g_{x+\boldsymbol{\delta}_m}-g_{x}$.
Considering that Hessian matrix $H_x$ is a real-valued matrix and can be decomposed as $H_x = V\Lambda V^{-1}$, where $\Lambda$ is a diagonal matrix: $\Lambda=\mathrm{diag}[\lambda_1, \lambda_2, ..., \lambda_r]$.
$V=\left[v_1, v_2, \cdots, v_n\right] \in \mathbb{R}^{n \times n}$  contains $n$ linearly independent eigenvectors, which means that $v_i^T * v_j = 0$, for $i \ne j$.
Without loss of generality, we normalize the eigenvectors, $V^TV=I$.
In this case, $H_x$ can be decomposed as $H_x=V \Lambda V^T$.
Therefore, it is possible to derive an easy-to-compute upper bound for $\Delta_g$,  as given by the following proposition.
\begin{proposition}\label{pro1}
Suppose that Hessian matrix $H_x=V \Lambda V^T$, where  $\Lambda=\mathrm{diag}[\lambda_1, \lambda_2, ..., \lambda_r]$ and $V=\left[v_1, v_2, \cdots, v_n\right]$. 
$\epsilon$ is the constraint of the $\ell_p$ norm of the adversarial perturbation, i.e., $||\boldsymbol{\delta}||_p \leq \epsilon$.
Then we have an upper bound of $||\Delta_g||$, given by:
\begin{equation}\label{eq:upperbound}
    ||\Delta_g|| \leq \sum_i |(e^{\epsilon \lambda_i}-1)| \cdot |g_{x}^Tv_i| \cdot ||v_i||
\end{equation}
\end{proposition}

Proposition~\ref{pro1} is proven in Appendix~\ref{appendix:prop1}, which establishes an important insight that although $\boldsymbol{\delta}$ has already produced a significant change in classification results, it is noteworthy that the upper-bound constraint on gradient changes in the saliency map can constrain its ability to capture the most class-discriminative information within images.
Consequently, the effectiveness of the gradient-based saliency map in revealing critical image information is limited under such constraints.
To address this limitation, we propose our CDIEA to identify the most class-discriminative information in images.


 \subsection{Class-Discriminative Information Estimation Algorithm (CDIEA)}~\label{sec:SAA}
 As can be seen in Fig.~\ref{framework}, CDIEA is designed to automatically estimate the critical points within images, along with their corresponding importance scores.
CDIEA is motivated by the realm of sparse adversarial attack research, which suggests that adding imperceptible perturbations to images can cause misclassification.
 Therefore, these pixels are also vulnerable to minor perturbations and similarly vulnerable to perturbations caused by data augmentation techniques.
 The importance scores indicate the extent to which these points influence classification decisions. 

Specifically, we first formulate the optimization goal of CDIEA as follows:
\begin{equation}\label{problem}
    \begin{array}{ll}
		\underset{{\boldsymbol{\delta}}}{\text{min}} & \|\boldsymbol{\delta}\|_{0} \\
		\text { s.t. } 	& \boldsymbol{\delta} \in[-\epsilon, \epsilon]^{w \times h \times c}\\
        &\underset{i=1,...,K}{\mathop{\arg\max}} f_i(\boldsymbol{x}+\boldsymbol{\delta}) \neq  y_{true} \\
	\end{array}
\end{equation}
where $\boldsymbol{\delta}$ is the adversarial perturbation, $\epsilon$ is the maximum perturbation magnitude, $f_i$ is a classification model and classifies $\boldsymbol{x}+\boldsymbol{\delta}$ as class $i$.
$\boldsymbol{x}$ and $y_{true}$ are the original image and its actual label, respectively.
The classification-sensitive information is typically distributed throughout the whole image~\cite{class_sense1,class_sense2}, extending beyond single sub-regions to encompass both foreground and background regions. 
Therefore, we firmly limit the magnitude of perturbation of $\epsilon$, to identify more critical points instead of just a few highly sensitive ones subject to significant disturbance.

To precisely obtain the positions of critical points, we encode the perturbation $\boldsymbol{\delta}$ through a neural network $ \mathcal{F}$: $\mathbb{R}^{w \times h \times c} \rightarrow \mathbb{R}^{w \times h \times 1}$. 
$\mathcal{F}(\boldsymbol{\delta})$ is of the same size as the image, with each element indicating the presence of a critical pixel.
Notably, the elements within the encoded tensor $\mathcal{F}(\boldsymbol{\delta})$ are inherently real numbers. 
To ensure continuity and differentiability, we adopt the scaled sigmoid function~\cite{autopruner} to generate an approximate binary matrix $\boldsymbol{m}=\mathrm{sigmoid}(\alpha \cdot \mathcal{F}(\boldsymbol{\delta}))$.
The scaling factor $\alpha$ controls the level of binarization.
For sufficiently large $\alpha$, $\boldsymbol{m}$ tends to approximate a binary form, and the approximate ones in $\boldsymbol{m}$ indicate the position of critical points.
During training, following the suggestions in~\cite{autopruner}, we gradually increase $\alpha$ from $\alpha_{init}$ to $\alpha_{end}$ to ensure accurate critical point estimation.
The degree of binarization of $\boldsymbol{m}$ is measured by:
\begin{equation}\label{eq:degree-of-binary}
 \mathcal{B}(\boldsymbol{m}) = \frac{1}{N}\left(\sum_{i=1}^N \mathbb{I}\left(\boldsymbol{m}_i<0.01\right)+\sum_{i=1}^N \mathbb{I}\left(\boldsymbol{m}_i>0.99\right)\right)  
\end{equation}
Therefore, when $\boldsymbol{m}_i$ falls below 0.01 or exceeds 0.99, it is deemed that $\boldsymbol{m}_i$ has effectively converged to either 0 or 1, respectively.
Meanwhile, the sparse perturbation $\boldsymbol{\delta} \odot \boldsymbol{m}$ ($\odot$ denotes element-wise multiplication) encloses the class-discriminative information, including both positions and varying degrees of importance.
A smaller value within the perturbation (denoted as $\boldsymbol{\delta}$) indicates that a lower level of perturbation is sufficient to achieve a significant impact.
Therefore, the corresponding pixel is more vulnerable to influence and thus highlights its increased importance.
 \begin{algorithm}[t]
\caption{The general workflow of CDIEA.}\label{alg:CDIEA}
\label{algorithm1}
\label{euclid}
\begin{algorithmic}[1]
\REQUIRE {an image $\boldsymbol{x}$, maximum number of iterations $T$, classification model $f$, maximum perturbation magnitude $\epsilon$, hyperparameter $v$, scaling factor $\alpha_{init}$ and $\alpha_{end}$, update step $\beta$}
\ENSURE adversarial perturbation $\boldsymbol{\delta}$
\STATE $\boldsymbol{\delta}_0$ $\leftarrow$ 0
\STATE $\boldsymbol{g}_0$ $\leftarrow$ 0
\STATE $\alpha_0$ $\leftarrow$ $\alpha_{init}$
\STATE $\alpha_{step}$ $\leftarrow$ $(\alpha_{end}-\alpha_{init})/T$
\STATE Randomly initialize the encoder $\mathcal{F}_0$
\FOR{$t=0$:$T-1$}
    \STATE $\boldsymbol{m}_t$ $\leftarrow$ sigmoid($\alpha_t \cdot \mathcal{F}_t(\boldsymbol{\delta}_t)$)
    \STATE Calculate the dynamic parameter $\mu_t$ based on $\boldsymbol{m}_t$ and Eq.~\eqref{eq:mu}
    \STATE Calculate the loss $\mathcal{L}$ according to $\boldsymbol{m}_t$, $\boldsymbol{\delta}_t$, $\mu_t$ and Eq.~\eqref{eq:loss-function}
    \STATE Update $\boldsymbol{g}_{t+1}$ and $\boldsymbol{\delta}_{t+1}$ according to Eq.~\eqref{eq:update}
    \STATE Update the encoder $\mathcal{F}_{t+1}$ based on the loss $\mathcal{L}$ and SGD with momentum
    \IF{$t/T >0.9$ and $ \mathcal{B}(\boldsymbol{m})<0.99$ }
    \STATE    $\alpha_{t+1}$ $\leftarrow$ $\alpha_{t} + 10 \cdot \alpha_{step}$
    \ELSE
    \STATE    $\alpha_{t+1} \leftarrow \alpha_{t} + \alpha_{step}$
    \ENDIF
\ENDFOR
\STATE $\boldsymbol{m}_T$ $\leftarrow$ sigmoid($\alpha_T \cdot \mathcal{F}_T(\boldsymbol{\delta}_T)$)
\STATE $\boldsymbol{\delta}$ $\leftarrow$ $\boldsymbol{\delta_T}\odot \boldsymbol{m}_T$
\RETURN $\boldsymbol{\delta}$
\end{algorithmic}
\end{algorithm}
Based on the aforementioned, we formulate our loss function as follows:
\begin{equation}\label{eq:loss-function}
\mathcal{L}=\mathcal{L}_{\mathrm{CE}}\left(f(\boldsymbol{x}+\boldsymbol{\delta} \odot \boldsymbol{m}), y^*\right)+\mu \frac{\|\boldsymbol{m}\|_1}{N}
\end{equation}
where $\|\cdot \|_1$, denotes the $\ell_1$-norm of the vector, $\mathcal{L}_{\mathrm{CE}}$ is the cross-entropy loss, and $y^* \neq y_{true}$.
Since $\boldsymbol{m}$ is approximately binarized, the second term approximately represents the sparsity of $\boldsymbol{m}$.
$\mu$ is a dynamic parameter to balance the two terms, which is calculated by:
\begin{equation}\label{eq:mu}
    \mu=\frac{\nu}{N} \sum_{i=1}^N \mathbb{I}\left(m_i>0.5\right)
\end{equation}
where $m_i$ is the $i$-th element of the mask $\boldsymbol{m}$, $\nu$ 
is a parameter to adjust the numeric disparity between loss items.
Considering the constrain of $\boldsymbol{\delta}$ in Eq.~\eqref{problem}, we utilize dense attack strategy~\cite{pgdl,fgsm} to quickly update $\boldsymbol{\delta}$ under the $\ell_\infty$-norm constrain.
In this way, we can constrain both the $\ell_0$-norm and $\ell_\infty$-norm of the perturbation. Specifically, the update formula of $\boldsymbol{\delta}$ is as follows:
\begin{equation}\label{eq:update}
	\begin{aligned}
		&\boldsymbol{g}_{t+1}=\sigma \cdot \boldsymbol{g}_{t}+\frac{\nabla_{\boldsymbol{x}} \mathcal{L}}{\left\|\nabla_{\boldsymbol{x}} \mathcal{L}\right\|_{1}} \\
		&\boldsymbol{\delta}_{t+1}=\operatorname{Clip}_{\boldsymbol{\epsilon}}\left\{\boldsymbol{\delta}_{t}-\beta \cdot \operatorname{sign}\left(\boldsymbol{g}_{t+1}\right)\right\}
	\end{aligned}
\end{equation}
where $\mathcal{L}$ is the loss defined by Eq.~\eqref{eq:loss-function}, $\sigma$ denotes the momentum decay factor, $\beta$ represents the update step, and $\operatorname{Clip}_{\boldsymbol{\epsilon}}\left\{ \cdot \right\}$ is used to project adversarial perturbation into the $\ell_\infty$-ball of radius $\epsilon$.
Both $\boldsymbol{g}_{0}$ and $\boldsymbol{\delta}_{0}$ are initialized to zero. Upon completing the iterative process, $\boldsymbol{m}$ will be strictly binarized with a threshold of 0.5 to indicate the positions of the critical points explicitly.
The general workflow of CDIEA is outlined in Algorithm~\ref{algorithm1}.

In particular, to minimize the computational cost induced by CDIEA and enhance the practicability of our framework, the generation of class-discriminative information is conducted offline.
Once this information is obtained for each dataset, the downstream model training utilizing IPF-RDA will no longer incorporate any training or inference from CDIEA.
This preserves the effectiveness of the framework with a streamlined and efficient implementation.

\subsection{Information-Preserving Framework (IPF)}~\label{sec:IPF}
Within this section, we describe the information-preserving framework and its integration with diverse data augmentation methods, as classified within the taxonomy depicted in Fig.~\ref{taxnomy}.
It is known that the efficacy of augmentation techniques benefits from providing augmented samples exhibiting greater diversity~\cite{evaluation}.
Guided by this notion, the core idea of our framework is to strike a balance between diversity enhancement and the preservation of class-discriminative information, which is a slight tradeoff.
Based on Eq.~\eqref{problem}, smaller perturbation values within $\boldsymbol{\delta}$ align with heightened sensitivity, where even minor perturbations significantly influence the pixels. 
Meanwhile, as suggested in previous works~\cite{gridmask,advmask, keepaugment}, which underscore the effectiveness of retaining structural regions, our framework focuses on preserving a structurally informative sub-region during each iteration.
Consequently, the importance map $\boldsymbol{\eta}$ of the critical pixels can be calculated as the reciprocal of the perturbation, i.e.,
\begin{equation}
        \boldsymbol{\eta}_{ij} =
    \begin{cases}
        1/ (\boldsymbol{\delta}_{ij} * \boldsymbol{m}_{ij}), & \text{if $\boldsymbol{m}_{ij} \neq 0$} \\
        0, & \text{otherwise}
     \end{cases}
\end{equation}
where $i=1,2,...,w$ and $j=1,2,...,h$.
The sensitivity of individual pixels is encapsulated in the non-zero entries of $\boldsymbol{\eta}$, serving as a direct indicator of their significance. 
Therefore, our framework introduces an importance score $\mathcal{IS}$ for rectangular regions $S$, which is defined as:
\begin{equation}\label{eq:is}
    \mathcal{IS}(\boldsymbol{\eta},S)=\sum_{(i,j) \in S} \boldsymbol{\eta}_{ij}.
\end{equation}
If $\mathcal{IS}(\boldsymbol{\eta}, S)$ attains a larger value, $S$ comprises more critical points with high importance scores.
These regions, thus, are deliberately safeguarded against any data augmentation-induced alterations.
Accurate and efficient assessments of the importance of arbitrary regions, represented by $S$, are crucial.
Addressing this, we undertake a statistical aggregation of $\mathcal{IS}$ values pertaining to all potential regions $S$ of a certain length $l$ offline.
The $\tau$-th percentile, donated as $Q_{\tau \mid \mathcal{IS}}$, serves as the discerning threshold.
This threshold signifies that a region $S$ exhibiting an $\mathcal{IS}$ value higher than $Q_{\tau \mid \mathcal{IS}}$ is more likely to contain critical information and should be carefully preserved.
As a result, the additional computational overhead introduced by our framework during training is solely to sample a region $S$ with an appropriate importance score.
This additional computational overhead is trivial when compared to the predominant costs intrinsic to model training efforts, indicating the high efficiency of our framework.

Based on the taxonomy in Fig.~\ref{taxnomy}, our proposed framework carefully preserves critical information to enhance the robustness of different data augmentation types.
\paragraph{Regional Deletion}
We prevent regional deletion-based data augmentation methods from inadvertently removing critical regions with high $\mathcal{IS}$.
To achieve this, we enforce a constraint on the importance score $\mathcal{IS}$ of randomly removed regions; the $\mathcal{IS}$ is rigorously below the threshold, denoted as $Q_{\tau \mid \mathcal{IS}}$.
For an image and the corresponding class-discriminative information $\boldsymbol{\eta}$, our framework ensures that the randomly removed region $S$ satisfies: $\mathcal{IS}(\boldsymbol{\eta}, S) \leq Q_{\tau \mid \mathcal{IS}}$.
In this way, critical information, such as objects of interest, remains almost unaffected during the augmentation process.
Our framework achieves enhanced robustness for regional deletion-based DA methods and concurrently maintains a diversified set of augmented data.
Striking this balance is of paramount importance, as it facilitates the protection of critical information while preserving the essential variability necessary for comprehensive and effective data augmentation strategies.

 \begin{algorithm}[t]
\caption{The pseudocode of IPF for robust data augmentation}
\label{alg:framework}
\begin{algorithmic}[1]
\REQUIRE {A batch of images $\boldsymbol{x}$, importance map $\boldsymbol{\eta}$, $\tau$-th percentile $Q_{\tau \mid \mathcal{\boldsymbol{\eta}}}$, batch size $N$, length $l$, DA method $\mathcal{A}$
}
\FOR{$i=0$:$N$}
\STATE \textit{(a) \textbf{Regional deletion}} 
    \REPEAT
        \STATE Sample a rectangle region $S$ of length $l$ in $\boldsymbol{\eta}_i$
        \STATE Compute $\mathcal{IS}$ of $S$ according to Eq.\eqref{eq:is}
   \UNTIL $\mathcal{IS}(\boldsymbol{\eta}_i, S) \leq Q_{\tau \mid \mathcal{IS}}$
   \STATE Erase region $S$ w.r.t. DA methods
   
\STATE \textit{(b) \textbf{Image-level transformation}}
  \REPEAT
     \STATE Sample a rectangle region $S$ of length $l$ in $\boldsymbol{\eta}_i$
     \STATE Compute $\mathcal{IS}$ of $S$ according to Eq.\eqref{eq:is}
  \UNTIL $\mathcal{IS}(\boldsymbol{\eta}_i, S) \geq Q_{\tau \mid \mathcal{IS}}$
  \STATE Recover region $S$ in the augmented image $\mathcal{A}(x_i)$
    
\STATE \textit{(c) \textbf{Image mixing}}
    \STATE Permute the current batch of images as $\boldsymbol{x}'$
    \REPEAT 
        \STATE Sample a rectangle region $S$ of length $l$ in $\boldsymbol{\eta}_i$
        Compute $\mathcal{IS}$ of $S$ according to Eq.\eqref{eq:is}
    \UNTIL $\mathcal{IS}(\boldsymbol{\eta}_i, S) \leq Q_{\tau \mid \mathcal{IS}}$
    \STATE Replace the region $S$ with a random region $S'$ in $\boldsymbol{x}'$ according to Eq.~\eqref{eq:image-mixing}
    \STATE Compute weight $\lambda$ to fuse labels according to Eq.~\eqref{eq:lambda}
    \ENDFOR
\ENSURE A batch of augmented images
\end{algorithmic}
\end{algorithm}

\paragraph{Image-level Transformation}
In the context of image-level augmentation, our framework aims to recover a random critical sub-region $S$ with $\mathcal{IS}$ higher than $Q_{\tau \mid \mathcal{IS}}$ after the augmentation process.
Meanwhile, the remainder of the image regions undergoes comprehensive augmentation via the employed data augmentation techniques.

The adoption of the proposed framework achieves dual objectives.
Primarily, it guarantees the diversity of the augmented data, which is crucial for practical training.
Simultaneously, the framework preserves classification-sensitive insights from perturbation during augmentation, thus bolstering the inherent robustness of the data augmentation approaches.

\paragraph{Image Mixing}
Our framework integrates region-level mixing data augmentation approaches through a two-fold approach to maintain the simplicity and efficiency of the original method pipeline.
To this end, we first prioritize the preservation of label information by substituting a less critical region in the first image $x_a$, thus preserving the information of label $y_a$ to the greatest extent possible.
Concurrently, we improve the calculation of label coefficients in Eq.~\eqref{eq:image-mixing} based on the ratio of the importance scores of the replaced regions $\mathbf{M}_b$ from the second image $x_b$ within the composite image $\tilde{x}$ as follows:
 \begin{equation}\label{eq:lambda}
     \omega = \frac{\mathcal{IS}(\mathbf{M}_b)}{\mathcal{IS}(\tilde{x})}.
 \end{equation}
 This modified strategy yields a logical advantage. 
 Specifically, when the chosen region from $x_b$ does not encompass any object associated with $y_b$, the $\mathcal{IS}(\mathbf{M}_b)$ becomes very small and negligible.
Consequently, the corresponding weight $\omega$ is substantially reduced, leading the fused label $\tilde{y}$ to be predominantly $y_a$. 
Therefore, this strategy yields a more coherent and rational way to minimize potential noise stemming from the data augmentation operations.
Finally, we present the pseudocode of IPF in Algorithm~\ref{alg:framework}.
\section{Experiments}
 This section empirically validates the proposed IPF-RDA across various tasks and investigates its behaviors through ablation studies conducted on seven widely used benchmark datasets.
We incorporate 10 commonly used SOTA DA methods into our framework to further enhance their robustness and practical performance.
Specifically, our evaluation encompasses a diverse range of tasks, commencing with supervised image classification on widely used benchmark datasets, including CIFAR-10/100~\cite{cifar-10} and Tiny-ImageNet~\cite{tiny}. 
Subsequently, the focus shifts to assessing the effectiveness of IPF-RDA within the context of person re-identification tasks to demonstrate the scalability of IPF-RDA.
This experiment is conducted on the CUHK03~\cite{cuhk03} and Market1501~\cite{market1501} datasets.
Furthermore, we extend the application of IPF-RDA to semi-supervised learning algorithms~\cite{uda},  thereby underscoring its versatility across multiple learning paradigms. 
Additionally, a thorough comparison of the effectiveness of IPF-RDA and KeepAugment~\cite{keepaugment} is conducted.

Besides, to gain deeper insights into our framework, we have meticulously orchestrated a series of analytical experiments, aiming at elucidating the functioning and efficacy of the proposed methodology.
Firstly, to demonstrate the effectiveness of our approach, we present visualizations showcasing the results of augmented data by various methods.
Meanwhile, to demonstrate the performance improvements achieved by our framework (e.g., models with enhanced feature extraction capabilities), we offer a set of t-SNE visualizations~\cite{tsne} on the MNIST dataset~\cite{mnist}.
These visualizations serve as a demonstrative testament to the performance enhancements instilled by our framework.
Secondly, to validate the capability of IPF-RDA in promoting models to learn better representations, we visualize the class activation maps (CAM)~\cite{cam}.
The CAM is obtained by models trained with IPF-RDA on the Oxford Flower dataset~\cite{flower}.
We then employ transfer learning by employing the models trained by IPF-RDA as backbones on CIFAR-100, followed by fine-tuning for CIFAR-10. 
Thirdly, empirical results are provided to evaluate the magnitude of the gradient change in Eq.~\eqref{lab3} relative to the original gradient.
Next, we evaluate the effectiveness of CDIEA in capturing the class-discriminative information.
Moreover, we conduct experiments related to additional training costs to verify the efficiency of our approach.
Lastly, we perform ablation studies, a systematic exploration to investigate the effect of different hyperparameters on the performance of IPF-RDA.
We report the average results over three independent runs with different random seeds.
\begin{table*}[]
	\centering
	\renewcommand\arraystretch{1.2}
		\caption{Error rates (\%) on CIFAR-10 and CIFAR-100 are summarized in this table across deep models. * means the published results in previous papers. The better results are \textbf{bold-faced}.  \textbf{FAA}: fast-AutoAugment. }
		\resizebox{.9\textwidth}{!}{
			\begin{tabular}{c|l|llllll}
				\toprule[1.5pt]
				Dataset & Method & ResNet18~\cite{resnet} & ResNet50~\cite{resnet}  &ResNet-110~\cite{resnet} &WRN-28-10~\cite{wrn}&Shake-2-32~\cite{2017Shake}&PyramidNet~\cite{shakedrop}\\ \hline
                \multirow{19}{*}{CIFAR-10~\cite{cifar}}& baseline&4.72*&4.34&4.55&4.48&5.10&2.80\\ \cline{2-8}
                &Cutout~\cite{cutout}&3.99*&4.19*&3.56&3.08*&3.04*&1.80\\
                &IPF-Cutout&\textbf{3.22} \scriptsize{$\downarrow$0.77}  &\textbf{3.27} \scriptsize{$\downarrow$0.92}&\textbf{3.36} \scriptsize{$\downarrow$0.20}&\textbf{2.70} \scriptsize{$\downarrow$0.38}&\textbf{2.73} \scriptsize{$\downarrow$0.31}&\textbf{1.68} \scriptsize{$\downarrow$0.12}\\ \cline{2-8} 
                &CutMix~\cite{cutmix}&3.36*&3.19*&3.51&3.07*&3.53&1.72\\
                &IPF-CutMix&\textbf{3.18} \scriptsize{$\downarrow$0.18}&\textbf{2.84} \scriptsize{$\downarrow$0.35}&\textbf{3.24} \scriptsize{$\downarrow$0.27}&\textbf{2.45} \scriptsize{$\downarrow$0.63}&\textbf{2.93} \scriptsize{$\downarrow$0.60}&\textbf{1.53} \scriptsize{$\downarrow$0.19} \\ \cline{2-8}
                &AutoAugment~\cite{autoaugment}&3.49*&3.41&3.34&2.99&2.70&1.70\\
                &IPF-AutoAugment&\textbf{2.98} \scriptsize{$\downarrow$0.51}&\textbf{3.13} \scriptsize{$\downarrow$0.28}&\textbf{3.03} \scriptsize{$\downarrow$0.31}&\textbf{2.57} \scriptsize{$\downarrow$0.42}&\textbf{2.37} \scriptsize{$\downarrow$0.33}&\textbf{1.31} \scriptsize{$\downarrow$0.39} \\ \cline{2-8}
                &RandAugment~\cite{randaugment}&3.53&3.75&4.03&3.06*&2.95&1.50\\
                &IPF-RandAugment &\textbf{3.21} \scriptsize{$\downarrow$0.32}&\textbf{2.82} \scriptsize{$\downarrow$0.93}&\textbf{3.29} \scriptsize{$\downarrow$0.74}&\textbf{2.48} \scriptsize{$\downarrow$0.58}&\textbf{2.57} \scriptsize{$\downarrow$0.38}&\textbf{1.38} \scriptsize{$\downarrow$0.12} \\ \cline{2-8}
                &Fast-AutoAugment~\cite{fast_autoaugment}&4.01&3.31&4.04&2.70*&3.58&1.58\\
                &IPF-FAA&\textbf{3.01} \scriptsize{$\downarrow$1.00}&\textbf{3.05} \scriptsize{$\downarrow$0.26}&\textbf{3.31} \scriptsize{$\downarrow$0.73}&\textbf{2.51} \scriptsize{$\downarrow$0.19}&\textbf{2.74} \scriptsize{$\downarrow$0.84}&\textbf{1.42} \scriptsize{$\downarrow$0.16} \\ \cline{2-8}
                &GridMask~\cite{gridmask}&3.62&3.85&4.03&2.77&3.09&1.69 \\
                &IPF-GridMask &\textbf{3.53} \scriptsize{$\downarrow$0.09}&\textbf{3.45} \scriptsize{$\downarrow$0.40}&\textbf{3.56} \scriptsize{$\downarrow$0.47}&\textbf{2.70} \scriptsize{$\downarrow$0.07}&\textbf{2.87} \scriptsize{$\downarrow$0.22}&\textbf{1.59} \scriptsize{$\downarrow$0.10}\\ \cline{2-8}
                &Has~\cite{has} &3.90&4.40&4.68&3.06&3.11*&2.05\\
                &IPF-HaS &\textbf{3.60} \scriptsize{$\downarrow$0.30}&\textbf{3.28} \scriptsize{$\downarrow$1.12}&\textbf{3.78} \scriptsize{$\downarrow$0.90}&\textbf{2.91} \scriptsize{$\downarrow$0.15}&\textbf{2.68} \scriptsize{$\downarrow$0.43}&\textbf{1.74} \scriptsize{$\downarrow$0.31}\\ \cline{2-8}
            &RandomErasing~\cite{random_erasing}&4.31&4.18&4.61&3.08&3.54*&2.21\\
                &IPF-RandomErasing&\textbf{3.63} \scriptsize{$\downarrow$0.68}&\textbf{3.33} \scriptsize{$\downarrow$0.85}&\textbf{3.71} \scriptsize{$\downarrow$0.90}&\textbf{2.72} \scriptsize{$\downarrow$0.36}&\textbf{2.81} \scriptsize{$\downarrow$0.73}&\textbf{1.77} \scriptsize{$\downarrow$0.44} \\ \cline{2-8}
                &AdvMask~\cite{advmask}&3.56*&3.31*&3.60&2.98*&2.97*&1.73 \\
                &IPF-AdvMask &\textbf{3.50} \scriptsize{$\downarrow$0.06}&\textbf{3.11} \scriptsize{$\downarrow$0.20}&\textbf{3.28} \scriptsize{$\downarrow$0.32}&\textbf{2.64} \scriptsize{$\downarrow$0.34}&\textbf{2.79} \scriptsize{$\downarrow$0.18}&\textbf{1.52} \scriptsize{$\downarrow$0.21} \\ \cline{2-8}
                &TrivialAugment~\cite{trivialaugment}&3.23&2.87&3.13&2.29&2.70&1.42\\
                &IPF-TrivialAugment&\textbf{3.18} \scriptsize{$\downarrow$0.05}&\textbf{2.85} \scriptsize{$\downarrow$0.02}&\textbf{3.04} \scriptsize{$\downarrow$0.09}&\textbf{2.23} \scriptsize{$\downarrow$0.06}&\textbf{2.63} \scriptsize{$\downarrow$0.07}&\textbf{1.34} \scriptsize{$\downarrow$0.08} \\ \bottomrule[1.5pt]
                &&ResNet-18~\cite{resnet}&ResNet-50~\cite{resnet}&ResNet-110~\cite{resnet}&WRN-28-10~\cite{wrn}&Shake-2-32~\cite{2017Shake}&Shake-2-96~\cite{2017Shake} \\ \hline
                \multirow{19}{*}{CIFAR-100~\cite{cifar}}&baseline&22.46*&22.59&23.83&21.04&23.35&18.34 \\ \cline{2-8}
                &Cutout~\cite{cutout} &21.96*&21.38&21.62&18.40*&20.63&16.54 \\
                &IPF-Cutout&\textbf{21.47} \scriptsize{$\downarrow$0.49}&\textbf{18.85} \scriptsize{$\downarrow$2.53}&\textbf{21.40} \scriptsize{$\downarrow$0.22}&\textbf{17.36} \scriptsize{$\downarrow$1.04}&\textbf{18.71} \scriptsize{$\downarrow$1.92}&\textbf{16.00} \scriptsize{$\downarrow$0.54} \\ \cline{2-8}
                &CutMix~\cite{cutmix} &19.55&18.66&19.32&17.33&20.43&15.29 \\
                &IPF-CutMix&\textbf{19.08} \scriptsize{$\downarrow$0.47}&\textbf{17.37} \scriptsize{$\downarrow$1.29}&\textbf{19.20} \scriptsize{$\downarrow$0.12}&\textbf{16.10} \scriptsize{$\downarrow$1.23}&\textbf{19.80} \scriptsize{$\downarrow$0.63}&\textbf{15.03} \scriptsize{$\downarrow$0.26}  \\ \cline{2-8}
                &AutoAugment~\cite{autoaugment} &20.22&17.92&20.37&17.10*&17.71&14.28 \\
                &IPF-AutoAugment&\textbf{19.62} \scriptsize{$\downarrow$0.60}&\textbf{17.50} \scriptsize{$\downarrow$0.42}&\textbf{18.11} \scriptsize{$\downarrow$2.26}&\textbf{16.24} \scriptsize{$\downarrow$0.86}&\textbf{17.38} \scriptsize{$\downarrow$0.33}&\textbf{14.08}  \scriptsize{$\downarrow$0.20}\\ \cline{2-8}
                &RandAugment~\cite{randaugment} &21.70&19.05&22.32&17.10*&20.00&17.02 \\
                &IPF-RandAugment&\textbf{20.94} \scriptsize{$\downarrow$0.76}&\textbf{18.10} \scriptsize{$\downarrow$0.95}&\textbf{20.95} \scriptsize{$\downarrow$1.37}&\textbf{16.38} \scriptsize{$\downarrow$0.72}&\textbf{19.26} \scriptsize{$\downarrow$0.74}&\textbf{16.60}  \scriptsize{$\downarrow$0.42}\\ \cline{2-8}
                &Fast-AutoAugment~\cite{fast_autoaugment} &20.89&20.92&23.87&17.30*&18.61&15.41 \\
                &IPF-FAA&\textbf{20.00} \scriptsize{$\downarrow$0.89}&\textbf{17.94} \scriptsize{$\downarrow$2.98}&\textbf{18.97} \scriptsize{$\downarrow$4.90}&\textbf{16.55} \scriptsize{$\downarrow$0.75}&\textbf{17.12} \scriptsize{$\downarrow$1.49}&\textbf{14.60} \scriptsize{$\downarrow$0.81} \\ \cline{2-8}
                &GridMask~\cite{gridmask} &24.77&21.62&21.27&19.60&20.57&17.45 \\
                &IPF-GridMask&\textbf{23.64} \scriptsize{$\downarrow$1.13}&\textbf{20.13} \scriptsize{$\downarrow$1.49}&\textbf{21.20} \scriptsize{$\downarrow$0.07}&\textbf{17.53} \scriptsize{$\downarrow$2.07}&\textbf{19.20} \scriptsize{$\downarrow$1.37}&\textbf{17.30} \scriptsize{$\downarrow$0.15} \\ \cline{2-8}
                &HaS~\cite{has} &21.81&21.24&22.11&19.78&21.03&15.92 \\
                &IPF-HaS&\textbf{21.63} \scriptsize{$\downarrow$0.18}&\textbf{18.63} \scriptsize{$\downarrow$2.61}&\textbf{20.85} \scriptsize{$\downarrow$1.26}&\textbf{17.57} \scriptsize{$\downarrow$2.21}&\textbf{19.76} \scriptsize{$\downarrow$1.27}&\textbf{15.84} \scriptsize{$\downarrow$0.08} \\ \cline{2-8}
                &RandomErasing~\cite{random_erasing} &24.03&22.21&22.85&19.43&21.19&18.36 \\
                &IPF-RandomErasing &\textbf{21.61} \scriptsize{$\downarrow$2.42}&\textbf{20.89} \scriptsize{$\downarrow$1.32}&\textbf{21.34} \scriptsize{$\downarrow$1.51}&\textbf{17.52} \scriptsize{$\downarrow$1.91}&\textbf{20.03} \scriptsize{$\downarrow$1.16}&\textbf{17.10} \scriptsize{$\downarrow$1.26} \\ \cline{2-8} 
                &AdvMask~\cite{advmask}&21.57*&21.01*&21.20&19.30*&20.04*&16.94 \\
                &IPF-AdvMask &\textbf{21.42} \scriptsize{$\downarrow$0.15}&\textbf{19.19} \scriptsize{$\downarrow$1.82}&\textbf{20.93} \scriptsize{$\downarrow$0.27}&\textbf{18.60} \scriptsize{$\downarrow$0.70}&\textbf{19.01} \scriptsize{$\downarrow$1.03}&\textbf{16.05} \scriptsize{$\downarrow$0.89} \\ \cline{2-8}
                &TrivialAugment~\cite{trivialaugment} &21.33&18.67&20.41&16.31&17.86&15.70 \\
                &IPF-TrivialAugment&\textbf{20.67} \scriptsize{$\downarrow$0.66}&\textbf{18.35} \scriptsize{$\downarrow$0.32}&\textbf{18.99} \scriptsize{$\downarrow$1.42}&\textbf{16.20} \scriptsize{$\downarrow$0.11}&\textbf{17.52} \scriptsize{$\downarrow$0.34}&\textbf{15.29} \scriptsize{$\downarrow$0.41} \\  
		\bottomrule[1.5pt]
		\end{tabular}}
		\label{cifar}
\end{table*}
\subsection{Experimental Setups for Image Classification}
We incorporate ten commonly used DA methods into our framework, including Cutout~\cite{cutout}, CutMix~\cite{cutmix}, RandomErasing~\cite{random_erasing}, HaS~\cite{has}, GridMask~\cite{gridmask}, AdvMask~\cite{advmask}, AutoAugment~\cite{autoaugment}, Fast-AutoAugment~\cite{fast_autoaugment}, RandAugment~\cite{randaugment}, and TrivialAugment~\cite{trivialaugment}.
In the context of supervised learning, we closely follow previous works~\cite{cutout,autoaugment,advmask,keepaugment} with our setup.
Specifically, all images are preprocessed by dividing each pixel value by 255 and normalizing by the dataset-specific statistics.
For the CIFAR dataset, we pad images to $36\times36$, randomly crop them into $32\times32$, and horizontally flip them with a probability of 50\%.
We train 1800 epochs with cosine learning rate decay for Shake-Shake~\cite{2017Shake} and Pyramid-Shakedrop~\cite{shakedrop} using SGD with Nesterov Momentum and a learning rate of 0.1, a batch size of 256, $1e^{-3}$ weight decay, and cosine learning rate decay. 
We conduct training for 300 epochs for all other deep networks, utilizing SGD with Nesterov Momentum and a learning rate of 0.1, a batch size of 128, a $5e^{-4}$ weight decay, and cosine learning rate decay.
Regarding the Tiny-ImageNet dataset, we resize the images into $64\times64$, initialize models with ImageNet pre-trained weights, and then fine-tune models using various augmentations.
The baseline approach involves solely padding and horizontal flipping.
To facilitate fair comparisons, all methods are implemented with the same training configurations. 

\textbf{Parameter Settings}
Following the suggestions in~\cite{autopruner}, we gradually increase $\alpha$ from $\alpha_{init}$ to $\alpha_{end}$ during training.
This adjustment serves a dual purpose: it facilitates the convergence of elements within $\boldsymbol{m}$ toward binary values while circumventing the algorithm's degeneration into random pixel selection.
 $\alpha_{end}$ is set to 100 on CIFAR and 10 on Tiny-ImageNet and others.
 The change of $\alpha_{init}$ has little effect on the results, so it is set to 0.1.
 $\gamma$ in Eq.~\eqref{eq:lambda} is typically set as 10 on CIFAR and 100 on other datasets.
 For IPF, we adopt the common settings from~\cite{cutout, keepaugment} with parameters $\tau$ and $l$ settings to minimize parameter tuning overhead.
 Specifically, $l$ is set to 16 on CIFAR-10 and Tiny-ImageNet, and 8 on CIFAR-100; $\tau$ is consistently fixed as 0.6 across datasets.
 The design of our framework ensures that these parameters are agnostic to the choice of network architectures and augmentation techniques.
Therefore, the parameters of our framework can be easily set, thus facilitating the seamless adoption and implementation of our approach.
\begin{table*}
    \centering
	\renewcommand\arraystretch{1.}
 \caption{ Error rates (\%) of different deep networks on the validation set of Tiny-ImageNet. * means the published results in previous papers. The better results are \textbf{bold-faced}. \textbf{FAA}: Fast-AutoAugment.}
        \small
    \resizebox{.9\textwidth}{!}{
			\begin{tabular}{l|lllllll}
				\toprule[1.5pt]
         Method & ResNet-18~\cite{resnet} & ResNet-50~\cite{resnet}  &WRN-50-2~\cite{wrn}&DenseNet121~\cite{densenet}&DenseNet201~\cite{densenet}&ResNext50~\cite{resnext}&ResNext101~\cite{resnext}\\ \midrule
        baseline & 38.62&26.39&18.45&37.99&31.08&20.24&14.94 \\ \midrule
        Cutout~\cite{cutout} &31.73&22.55&17.73&33.54&30.68&17.84&14.47 \\ 
        IPF-Cutout &\textbf{28.94} \scriptsize{$\downarrow$2.79}&\textbf{20.97} \scriptsize{$\downarrow$1.58} &\textbf{16.75} \scriptsize{$\downarrow$0.98}&\textbf{33.24} \scriptsize{$\downarrow$0.30}&\textbf{30.60} \scriptsize{$\downarrow$0.08}&\textbf{17.45} \scriptsize{$\downarrow$0.39}&\textbf{14.25} \scriptsize{$\downarrow$0.22} \\ \midrule
        CutMix~\cite{cutmix}&35.91&23.59&17.68&37.40&33.70&17.79&14.36 \\ 
        IPF-CutMix&\textbf{30.51} \scriptsize{$\downarrow$5.40}&\textbf{20.88} \scriptsize{$\downarrow$2.71}&\textbf{16.96} \scriptsize{$\downarrow$0.72}&\textbf{33.46} \scriptsize{$\downarrow$3.94}&\textbf{31.15} \scriptsize{$\downarrow$2.55}&\textbf{17.38} \scriptsize{$\downarrow$0.41}&\textbf{14.05} \scriptsize{$\downarrow$0.31} \\ \midrule
        RandomErasing~\cite{random_erasing} &36.00&24.67&18.11&34.38&31.68&18.48& 14.27\\ 
        IPF-RandomErasing &\textbf{28.58} \scriptsize{$\downarrow$7.42}&\textbf{20.89} \scriptsize{$\downarrow$3.78}&\textbf{16.73} \scriptsize{$\downarrow$1.38}&\textbf{33.38} \scriptsize{$\downarrow$1.00}&\textbf{31.01} \scriptsize{$\downarrow$0.67}&\textbf{17.43} \scriptsize{$\downarrow$1.05}&\textbf{14.07} \scriptsize{$\downarrow$0.20} \\ \hline
        TrivialAugment~\cite{trivialaugment} &29.98&24.40&16.54&38.86&31.10&19.09&14.23 \\ 
        IPF-TrivialAugment &\textbf{27.66} \scriptsize{$\downarrow$2.32}&\textbf{19.86} \scriptsize{$\downarrow$4.54}&\textbf{16.10} \scriptsize{$\downarrow$0.44}&\textbf{33.91} \scriptsize{$\downarrow$4.95}&\textbf{30.43} \scriptsize{$\downarrow$0.67}&\textbf{18.79} \scriptsize{$\downarrow$0.30}&\textbf{13.61} \scriptsize{$\downarrow$0.62} \\ \midrule
        AutoAugment~\cite{autoaugment}&31.72&29.75&17.48&34.76&32.83&18.72&15.06 \\ 
        IPF-AutoAugment&\textbf{30.54} \scriptsize{$\downarrow$1.18}&\textbf{23.42} \scriptsize{$\downarrow$6.33}&\textbf{17.01} \scriptsize{$\downarrow$0.47}&\textbf{34.43} \scriptsize{$\downarrow$0.33}&\textbf{32.29} \scriptsize{$\downarrow$0.54}&\textbf{18.41} \scriptsize{$\downarrow$0.31}&\textbf{14.40} \scriptsize{$\downarrow$0.66} \\ \midrule
        Fast-AutoAugment~\cite{fast_autoaugment}&32.26&21.06&17.16&33.65&31.70&18.96&14.78 \\ 
        IPF-FAA&\textbf{29.83} \scriptsize{$\downarrow$2.43}&\textbf{20.31} \scriptsize{$\downarrow$0.75}&\textbf{16.98} \scriptsize{$\downarrow$0.18}&\textbf{33.26} \scriptsize{$\downarrow$0.39}&\textbf{31.57} \scriptsize{$\downarrow$0.13}&\textbf{17.11} \scriptsize{$\downarrow$1.85}&\textbf{14.34} \scriptsize{$\downarrow$0.44} \\ \midrule
        GridMask~\cite{gridmask}&37.28*&22.12&17.75&33.99&31.36&18.95&14.12 \\ 
        IPF-GridMask&\textbf{30.78} \scriptsize{$\downarrow$6.50}&\textbf{21.52} \scriptsize{$\downarrow$0.60}&\textbf{16.49} \scriptsize{$\downarrow$1.26}&\textbf{33.51} \scriptsize{$\downarrow$0.48}&\textbf{31.24} \scriptsize{$\downarrow$0.12}&\textbf{17.45} \scriptsize{$\downarrow$1.50}&\textbf{13.64} \scriptsize{$\downarrow$0.48} \\ \midrule
        HaS~\cite{has}&36.49*&24.68&18.23&34.26&31.74&19.48&13.74 \\ 
        IPF-HaS&\textbf{36.27} \scriptsize{$\downarrow$0.22}&\textbf{23.89} \scriptsize{$\downarrow$0.79}&\textbf{17.81} \scriptsize{$\downarrow$0.42}&\textbf{33.90} \scriptsize{$\downarrow$0.36}&\textbf{31.60} \scriptsize{$\downarrow$0.14}&\textbf{18.56} \scriptsize{$\downarrow$0.92}&\textbf{13.26} \scriptsize{$\downarrow$0.48} \\ \midrule
        AdvMask~\cite{advmask}&34.71*&20.16&16.61*&33.84&30.92&17.91&14.17 \\
        IPF-AdvMask&\textbf{33.09} \scriptsize{$\downarrow$1.62}&\textbf{19.93} \scriptsize{$\downarrow$0.23}&\textbf{15.99} \scriptsize{$\downarrow$0.62}&\textbf{33.69} \scriptsize{$\downarrow$0.15}&\textbf{30.57} \scriptsize{$\downarrow$0.35}&\textbf{17.60} \scriptsize{$\downarrow$0.31}&\textbf{13.74} \scriptsize{$\downarrow$0.43} \\ \midrule
        RandAugment~\cite{randaugment}&34.33&24.13&17.75&36.05&32.25&19.64&14.80 \\ 
        IPF-RandAugment&\textbf{34.23} \scriptsize{$\downarrow$0.10}&\textbf{23.82} \scriptsize{$\downarrow$0.31}&\textbf{17.63} \scriptsize{$\downarrow$0.12}&\textbf{35.59} \scriptsize{$\downarrow$0.46}&\textbf{31.99} \scriptsize{$\downarrow$0.26}&\textbf{19.57}  \scriptsize{$\downarrow$0.07}&\textbf{14.73} \scriptsize{$\downarrow$0.07} \\ 
    \bottomrule[1.5pt]
    \end{tabular}}
    \label{tiny-imagenet}
\end{table*}
\subsection{Supervised Image Classification}
\subsubsection{Results on CIFAR-10 and CIFAR-100}
We conduct a comprehensive evaluation of our framework on the CIFAR-10 and CIFAR-100 datasets.
For evaluation, various popular deep networks are utilized, including ResNet-18/50/110~\cite{resnet}, Wide-ResNet-28-10 (WRN-28-10)~\cite{wrn}, ShakeShake2-32/96~\cite{2017Shake} and Pyramid-Shakedrop~\cite{shakedrop} whose hyper-parameters are precisely the same as those used in~\cite{fast_autoaugment, cutout, trivialaugment}.
The error rates on CIFAR-10/100 test sets are shown in Table~\ref{cifar}.

It can be observed that the original data augmentation techniques contribute to the improvements in model performance.
However, our proposed framework further enhances the efficacy of these SOTA data augmentation approaches.
Specifically, IPF-RDA consistently yields substantial improvements across nearly all data augmentation methods.
Particularly, on both CIFAR-10 and CIFAR-100, our method can achieve 0.1-1.0\% and 0.1-2.7\% error rate reductions, respectively.
For instance, after integrating Cutout into our framework, IPF-Cutout reduces the test error rates of the original Cutout using ResNet-50 on CIFAR-10 by 0.92 percent.
On CIFAR-100, for relatively small models like ResNet-18 and ResNet-50, IPF-RDA improves data augmentation approaches by about 0.8 percent.
In contrast, for larger models like ResNet-110, Wide-ResNet-28-10, and Shakeshake models, our framework can improve most methods by over 1 percent.

Meanwhile, our framework's effectiveness is particularly pronounced for information deletion-based data augmentation methods, such as Cutout, as well as for automatic augmentations, such as AutoAugment.
These methodologies, prone to introducing perturbations in augmented images, significantly benefit from the mitigating influence of IPF-RDA, manifesting substantial gains in overall performance.
A plausible explanation for the superior performance of our method is that our method suppresses noises caused by data augmentation and helps models to learn better generalization abilities. 

\subsubsection{Results on Tiny-ImageNet}
The comparative evaluation of augmentation methods, before and after their integration into our proposed framework, is presented in Table~\ref{tiny-imagenet} using the Tiny-ImageNet dataset.
For a comprehensive assessment, we employ a spectrum of popular deep network architectures, encompassing ResNet-18/50/110~\cite{resnet}, Wide-ResNet-50-2~\cite{wrn}, DenseNet121/201~\cite{densenet}, and ResNext50/101~\cite{resnext}.

It can be observed that the original data augmentation approaches contribute to performance improvements for deep models compared to the baseline model.
More importantly, the performance of IPF-RDA becomes unequivocally apparent, as it substantially elevates the overall generalization capabilities of these data augmentation strategies, effectively enhancing their robustness and unleashing their full potential during the model training process.
For instance, the error rates of ResNet-50, as witnessed in conjunction with the original augmentation methods, experience a reduction exceeding 1\% following the application of IPF-RDA.
Meanwhile, the extent of improvement can be discernibly influenced by the choice of architecture. 
Specifically, compared to ResNets and Wide-ResNets, DenseNets generally suffer less from overfitting due to the architecture design and thus appear to benefit less from data augmentation techniques.

\subsection{Person Re-Identification}
 \begin{table}[]
     \centering
     \caption{Detection mean Average Precision (mAP) Results on CUHK03~\cite{cuhk03} and Market1501~\cite{market1501} datasets. mAP (\%) is reported for comparison. }
     \label{table:ReID}
     \begin{tabular}{l|ll}
     \toprule[1.5pt]
         Method & CUHK03~\cite{cuhk03} & Market1501~\cite{market1501} \\ \hline
        Cutout~\cite{cutout} &81.9 & 89.8 \\
        IPF-Cutout &\textbf{82.1} \scriptsize{$\uparrow$0.2}  & \textbf{90.1} \scriptsize{$\uparrow$0.3} \\ \hline
        HaS~\cite{has} &77.1 & 86.9\\
        IPF-HaS &\textbf{77.5} \scriptsize{$\uparrow$0.4} & \textbf{88.4} \scriptsize{$\uparrow$1.5}\\ \hline
        RE~\cite{random_erasing} &81.6& 90.6 \\
        IPF-RE &\textbf{82.3} \scriptsize{$\uparrow$0.7}&\textbf{90.8} \scriptsize{$\uparrow$0.2} \\
        \bottomrule[1.5pt]
     \end{tabular}
 \end{table}
 Data augmentation is also widely used in person re-identification tasks.
To demonstrate the scalability of IPF-RDA, we apply the proposed algorithm in the context of the ReID task, specifically employing the CUHK03~\cite{cuhk03} and Market1501~\cite{market1501} datasets.

Since Cutout, HaS, and Random Erasing are the most commonly used data augmentation methods in ReID, we incorporate these methods into our framework.
We adopt the feature pyramid branch model~\cite{FPB}, using the standard hyperparameters from the codebase~\footnote{https://github.com/anocodetest1/FPB/}.
We fix nearly all the training setups but utilize various data augmentation approaches to assess their performance during training.
The mean average precision (mAP) scores across three independent trials are shown in Table~\ref{table:ReID}.
Evidently, it can be observed that IPF-RDA further improves the generalization performance of these methods significantly.
Specifically, when IPF-RDA is applied in conjunction with HaS, our framework outperforms the original HaS by 1.5\% and 0.4\% on Market1501 and CUHK03, respectively.
The reason behind the superior performance of our method lies in its ability to effectively suppress the complete or substantial removal of critical objects in images when applied alongside these methods, all while preserving data diversity. 
Therefore, models can be trained more effectively.
This emphasizes the efficacy and scalability of IPF-RDA in bolstering the performance of data augmentation strategies within the realm of ReID tasks.

 \begin{table}[]
     \centering
     \caption{Semi-supervised learning test accuracy on CIFAR-10.}
     \label{table:ssl}
     \begin{tabular}{l|ll}
     \toprule[1.5pt]
          Method& Acc (\%) & Prec (\%)  \\ \hline
          Cutout~\cite{cutout} &91.54 &91.01 \\
          IPF-Cutout &\textbf{91.71} \scriptsize{$\uparrow$0.17} & \textbf{91.70} \scriptsize{$\uparrow$0.69} \\ \hline
          HaS~\cite{has} &86.96 & 86.18\\
          IPF-HaS&\textbf{91.32} \scriptsize{$\uparrow$4.36} & \textbf{89.59} \scriptsize{$\uparrow$3.41}\\ \hline
          RandomErasing~\cite{random_erasing} &91.10&90.79 \\ 
          IPF-RE &\textbf{92.22} \scriptsize{$\uparrow$1.12} &\textbf{92.17} \scriptsize{$\uparrow$1.38} \\ \hline
          AutoAugment~\cite{autoaugment} &92.97&92.91 \\
          IPF-AA &\textbf{93.66} \scriptsize{$\uparrow$0.69}&\textbf{93.44} \scriptsize{$\uparrow$0.53}\\ \hline
          RandAugment~\cite{randaugment} &91.03 & 90.89 \\
          IPF-RA &\textbf{93.26} \scriptsize{$\uparrow$2.23} & \textbf{93.03} \scriptsize{$\uparrow$2.14} \\ \hline
          FAA~\cite{fast_autoaugment} &91.01 &91.11 \\
          IPF-FAA & \textbf{91.07} \scriptsize{$\uparrow$0.06}& \textbf{91.30} \scriptsize{$\uparrow$0.19} \\ \hline
          TrivialAugment~\cite{trivialaugment} & 93.34& 93.10 \\
          IPF-TA &\textbf{93.47} \scriptsize{$\uparrow$0.13} & \textbf{93.20} \scriptsize{$\uparrow$0.10} \\
        \bottomrule[1.5pt]
     \end{tabular}
 \end{table}
\subsection{Semi-Supervised Learning}\label{sec:ssl}
Data augmentation has been employed as a powerful tool for developing SOTA semi-supervised learning (SSL) algorithms. 
In light of this, we extend the application of the proposed framework to the domain of unsupervised data augmentation~\cite{uda} (UDA) on CIFAR-10, aiming to demonstrate the wide-ranging applicability of our method.
To facilitate our evaluation, we divide the training set of CIFAR-10 into two subsets: a labeled set comprising 4000 samples and an unlabeled set with others by randomly removing parts of the labels.
All training setups closely follow previous work~\cite{uda,ssl} but employ different data augmentations.

The results of both accuracy and precision metrics using Wide-ResNet-28-2~\cite{wrn} are shown in Table~\ref{table:ssl}. 
It can be observed that IPF-RDA consistently elevates the performance of diverse augmentations in terms of both accuracy and precision. 
Particularly noteworthy are the discernible improvements achieved by IPF-RDA, with substantial 4.36\%, 1.12\%, and 0.69\% improvements in test accuracy for HaS, RandomErasing, and AutoAugment, respectively. 
These findings conclusively underscore the efficacy of IPF-RDA within the realm of SSL tasks.
\newcommand{\lenn}{.4}
\newcommand{\Lenn}{0.15}
\newcommand{\vlenn}{-1.5}
\newcommand{\hlenn}{-1.8}
\newcommand{\hlennn}{-1.6}
\newcommand{\len}{1.}
\newcommand{\Len}{0.15}
\newcommand{\vlen}{-3}
\begin{figure*}[]
 \begin{flushleft}
		\footnotesize \hspace{1.5cm}Cutout \hspace{0.5mm} IPF-Cutout \hspace{3mm} HaS \hspace{3mm}IPF-HaS \hspace{1mm} GridMask  IPF-GM \hspace{5mm} AA \hspace{3mm} IPF-AA \hspace{5mm} RA \hspace{4mm} IPF-RA \hspace{2mm} CutMix \hspace{1mm} IPF-CM
	\end{flushleft}
	\centering
	\subfloat{
		\begin{minipage}{\Lenn \linewidth}
			\centering
			\setlength{\abovecaptionskip}{-0.01cm} 
			\includegraphics[width=\lenn \linewidth]{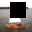}
		\end{minipage}
		\hspace{\hlenn cm}
		\begin{minipage}{\Lenn \linewidth}
			\centering
			\setlength{\abovecaptionskip}{-0.01cm} 
			\includegraphics[width=\lenn \linewidth]{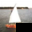}
		\end{minipage}
            \hspace{\hlennn cm}
    		\begin{minipage}{\Lenn \linewidth}
    			\centering
    			\setlength{\abovecaptionskip}{-0.01cm} 
    			\includegraphics[width=\lenn \linewidth]{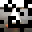}
    		\end{minipage}
            \hspace{\hlenn cm}
    		\begin{minipage}{\Lenn \linewidth}
    			\centering
    			\setlength{\abovecaptionskip}{-0.01cm} 
    			\includegraphics[width=\lenn \linewidth]{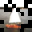}
    		\end{minipage}
            \hspace{\hlennn cm}
    		\begin{minipage}{\Lenn \linewidth}
    			\centering
    			\setlength{\abovecaptionskip}{-0.01cm} 
    			\includegraphics[width=\lenn \linewidth]{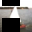}
    		\end{minipage}
            \hspace{\hlenn cm}
    		\begin{minipage}{\Lenn \linewidth}
    			\centering
    			\setlength{\abovecaptionskip}{-0.01cm} 
    			\includegraphics[width=\lenn \linewidth]{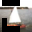}
    		\end{minipage}
            \hspace{\hlennn cm}
    		\begin{minipage}{\Lenn \linewidth}
    			\centering
    			\setlength{\abovecaptionskip}{-0.01cm} 
    			\includegraphics[width=\lenn \linewidth]{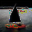}
    		\end{minipage}
            \hspace{\hlenn cm}
    		\begin{minipage}{\Lenn \linewidth}
    			\centering
    			\setlength{\abovecaptionskip}{-0.01cm} 
    			\includegraphics[width=\lenn \linewidth]{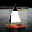}
    		\end{minipage}
            \hspace{\hlennn cm}
    		\begin{minipage}{\Lenn \linewidth}
    			\centering
    			\setlength{\abovecaptionskip}{-0.01cm} 
    			\includegraphics[width=\lenn \linewidth]{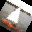}
    		\end{minipage}
            \hspace{\hlenn cm}
    		\begin{minipage}{\Lenn \linewidth}
    			\centering
    			\setlength{\abovecaptionskip}{-0.01cm} 
    			\includegraphics[width=\lenn \linewidth]{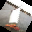}
    		\end{minipage}
             \hspace{\hlennn cm}
    		\begin{minipage}{\Lenn \linewidth}
    			\centering
    			\setlength{\abovecaptionskip}{-0.01cm} 
    			\includegraphics[width=\lenn \linewidth]{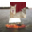}
    		\end{minipage}
            \hspace{\hlenn cm}
    		\begin{minipage}{\Lenn \linewidth}
    			\centering
    			\setlength{\abovecaptionskip}{-0.01cm} 
    			\includegraphics[width=\lenn \linewidth]{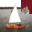}
    		\end{minipage}
            
	}
 \vspace{\vlen mm }
	\newline
	\subfloat{
		\begin{minipage}{\Lenn \linewidth}
			\centering
			\setlength{\abovecaptionskip}{-0.01cm} 
			\includegraphics[width=\lenn \linewidth]{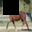}
		\end{minipage}
		\hspace{\hlenn cm}
		\begin{minipage}{\Lenn \linewidth}
			\centering
			\setlength{\abovecaptionskip}{-0.01cm} 
			\includegraphics[width=\lenn \linewidth]{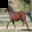}
		\end{minipage}
            \hspace{\hlennn cm}
    		\begin{minipage}{\Lenn \linewidth}
    			\centering
    			\setlength{\abovecaptionskip}{-0.01cm} 
    			\includegraphics[width=\lenn \linewidth]{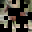}
    		\end{minipage}
            \hspace{\hlenn cm}
    		\begin{minipage}{\Lenn \linewidth}
    			\centering
    			\setlength{\abovecaptionskip}{-0.01cm} 
    			\includegraphics[width=\lenn \linewidth]{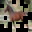}
    		\end{minipage}
            \hspace{\hlennn cm}
    		\begin{minipage}{\Lenn \linewidth}
    			\centering
    			\setlength{\abovecaptionskip}{-0.01cm} 
    			\includegraphics[width=\lenn \linewidth]{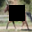}
    		\end{minipage}
            \hspace{\hlenn cm}
    		\begin{minipage}{\Lenn \linewidth}
    			\centering
    			\setlength{\abovecaptionskip}{-0.01cm} 
    			\includegraphics[width=\lenn \linewidth]{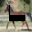}
    		\end{minipage}
            \hspace{\hlennn cm}
    		\begin{minipage}{\Lenn \linewidth}
    			\centering
    			\setlength{\abovecaptionskip}{-0.01cm} 
    			\includegraphics[width=\lenn \linewidth]{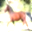}
    		\end{minipage}
            \hspace{\hlenn cm}
    		\begin{minipage}{\Lenn \linewidth}
    			\centering
    			\setlength{\abovecaptionskip}{-0.01cm} 
    			\includegraphics[width=\lenn \linewidth]{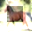}
    		\end{minipage}
            \hspace{\hlennn cm}
    		\begin{minipage}{\Lenn \linewidth}
    			\centering
    			\setlength{\abovecaptionskip}{-0.01cm} 
    			\includegraphics[width=\lenn \linewidth]{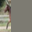}
    		\end{minipage}
            \hspace{\hlenn cm}
    		\begin{minipage}{\Lenn \linewidth}
    			\centering
    			\setlength{\abovecaptionskip}{-0.01cm} 
    			\includegraphics[width=\lenn \linewidth]{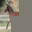}
    		\end{minipage}
             \hspace{\hlennn cm}
    		\begin{minipage}{\Lenn \linewidth}
    			\centering
    			\setlength{\abovecaptionskip}{-0.01cm} 
    			\includegraphics[width=\lenn \linewidth]{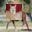}
    		\end{minipage}
            \hspace{\hlenn cm}
    		\begin{minipage}{\Lenn \linewidth}
    			\centering
    			\setlength{\abovecaptionskip}{-0.01cm} 
    			\includegraphics[width=\lenn \linewidth]{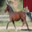}
    		\end{minipage}
	}
 \vspace{\vlen mm }
	\newline
	\subfloat{
            \hspace{-5.5 mm}
		\begin{minipage}{\Lenn \linewidth}
			\centering
			\setlength{\abovecaptionskip}{-0.01cm} 
			\includegraphics[width=\lenn \linewidth]{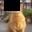}
		\end{minipage}
		\hspace{\hlenn cm}
		\begin{minipage}{\Lenn \linewidth}
			\centering
			\setlength{\abovecaptionskip}{-0.01cm} 
			\includegraphics[width=\lenn \linewidth]{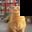}
		\end{minipage}
            \hspace{\hlennn cm}
    		\begin{minipage}{\Lenn \linewidth}
    			\centering
    			\setlength{\abovecaptionskip}{-0.01cm} 
    			\includegraphics[width=\lenn \linewidth]{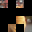}
    		\end{minipage}
            \hspace{\hlenn cm}
    		\begin{minipage}{\Lenn \linewidth}
    			\centering
    			\setlength{\abovecaptionskip}{-0.01cm} 
    			\includegraphics[width=\lenn \linewidth]{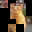}
    		\end{minipage}
            \hspace{\hlennn cm}
    		\begin{minipage}{\Lenn \linewidth}
    			\centering
    			\setlength{\abovecaptionskip}{-0.01cm} 
    			\includegraphics[width=\lenn \linewidth]{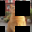}
    		\end{minipage}
            \hspace{\hlenn cm}
    		\begin{minipage}{\Lenn \linewidth}
    			\centering
    			\setlength{\abovecaptionskip}{-0.01cm} 
    			\includegraphics[width=\lenn \linewidth]{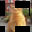}
    		\end{minipage}
            \hspace{\hlennn cm}
    		\begin{minipage}{\Lenn \linewidth}
    			\centering
    			\setlength{\abovecaptionskip}{-0.01cm} 
    			\includegraphics[width=\lenn \linewidth]{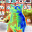}
    		\end{minipage}
            \hspace{\hlenn cm}
    		\begin{minipage}{\Lenn \linewidth}
    			\centering
    			\setlength{\abovecaptionskip}{-0.01cm} 
    			\includegraphics[width=\lenn \linewidth]{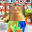}
    		\end{minipage}
            \hspace{\hlennn cm}
    		\begin{minipage}{\Lenn \linewidth}
    			\centering
    			\setlength{\abovecaptionskip}{-0.01cm} 
    			\includegraphics[width=\lenn \linewidth]{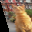}
    		\end{minipage}
            \hspace{\hlenn cm}
    		\begin{minipage}{\Lenn \linewidth}
    			\centering
    			\setlength{\abovecaptionskip}{-0.01cm} 
    			\includegraphics[width=\lenn \linewidth]{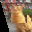}
    		\end{minipage}
             \hspace{\hlennn cm}
    		\begin{minipage}{\Lenn \linewidth}
    			\centering
    			\setlength{\abovecaptionskip}{-0.01cm} 
    			\includegraphics[width=\lenn \linewidth]{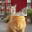}
    		\end{minipage}
            \hspace{\hlenn cm}
    		\begin{minipage}{\Lenn \linewidth}
    			\centering
    			\setlength{\abovecaptionskip}{-0.01cm} 
    			\includegraphics[width=\lenn \linewidth]{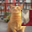}
    		\end{minipage}
	}
 \caption{Visualization of augmented data on CIFAR-10 using different DA approaches.}
 \label{vis-of-augmented-data}
 \vspace{-5mm}
\end{figure*}
 \begin{figure*}[]
	\centering
 \subfloat[ResNet-50, FAA, \\
            Test Acc: 96.06\% \\
            DI: $2.04\times 10^{-6}$
 ]{
		\includegraphics[width=0.5\columnwidth]{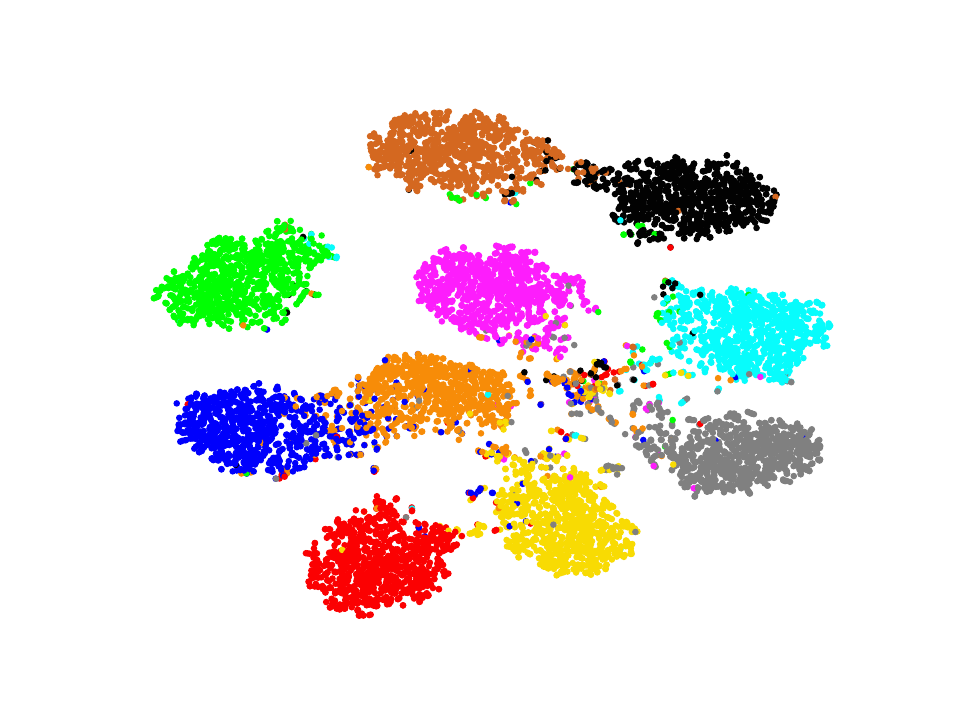}
        \label{faa}
	}  
	\subfloat[ResNet-50, IPF-FAA, \\
                Test Acc: 96.95\% \\
                DI: \text{$3.28\times 10^{-5}$}
    ]{
		\includegraphics[width=0.5\columnwidth]{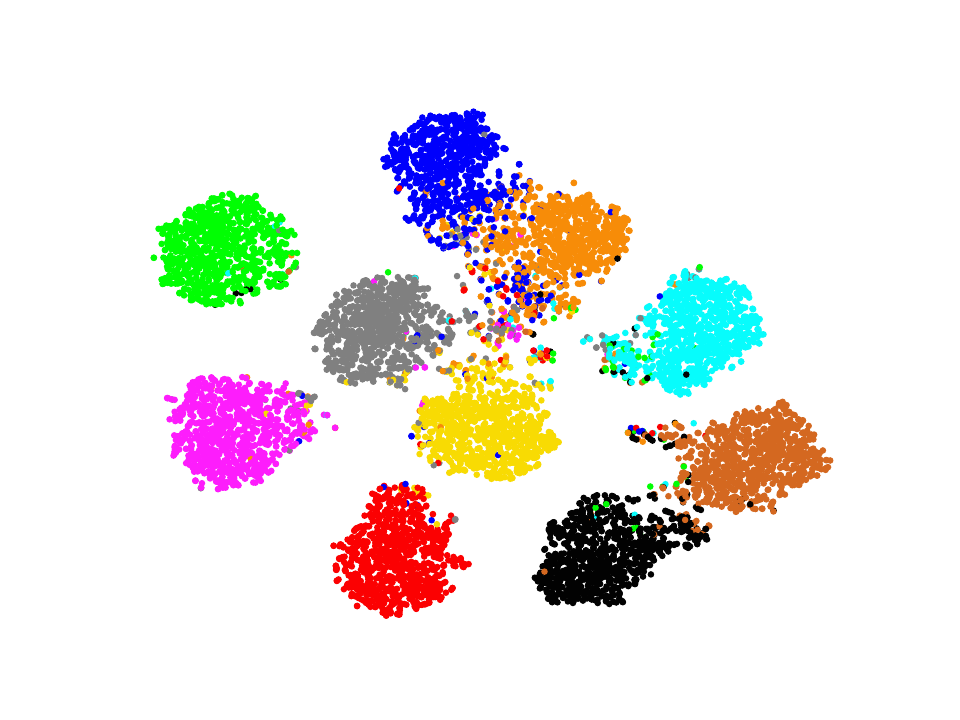}
        \label{ipf-faa}
	} 
	\subfloat[Transferred ResNet-50, \\
                FAA, Test Acc: 92.26\% \\
                    DI: $4.43\times 10^{-5}$
                ]{
		\includegraphics[width=0.5\columnwidth]{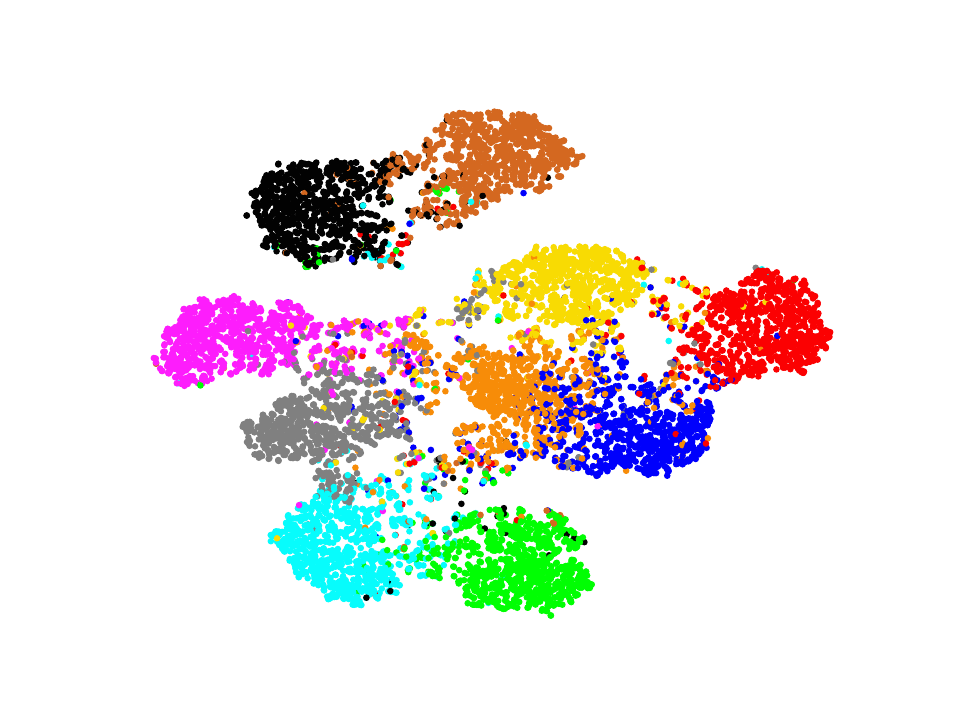}
        \label{trans-faa}
	} 	
     \subfloat[Transferred ResNet-50,\\
     IPF-FAA, Test Acc: 93.45\%\\
                    DI: $7.58\times 10^{-5}$
                ]{
		\includegraphics[width=0.5\columnwidth]{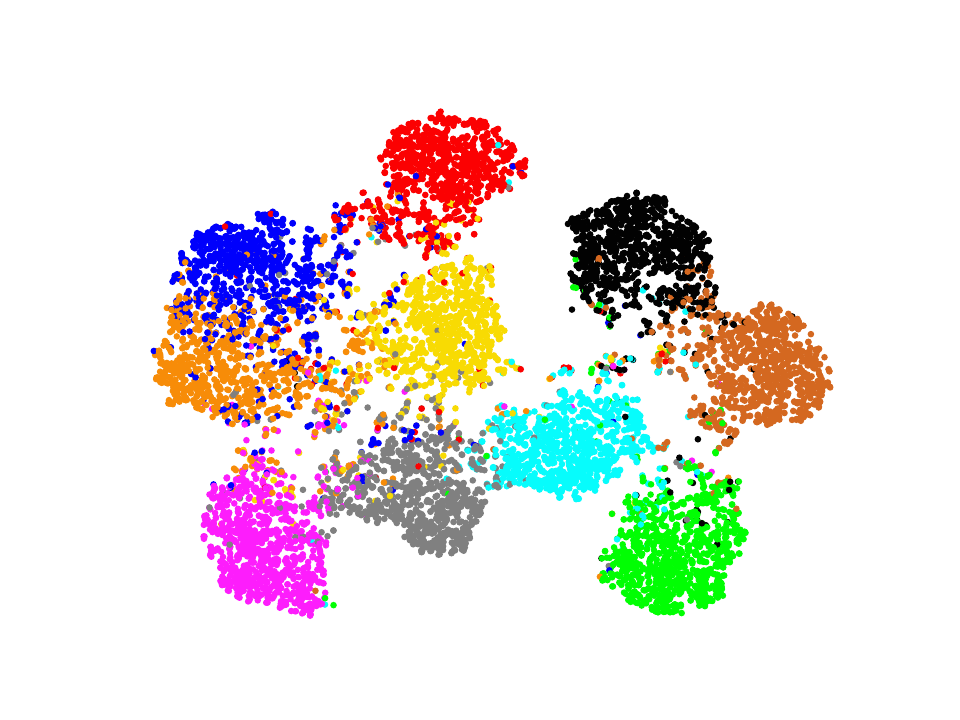}
    \label{trans-ipf-faa}
	} 	
	\caption{Visualization of deep features on CIFAR-10 using t-SNE algorithm~\cite{tsne}. Each color denotes a class. DI: Dunn index.}
	\label{fig:tsne}
\end{figure*}
  \begin{table}[]
     \centering
     \caption{Comparison between IPF-RDA and KeepAugment based on Cutout~\cite{cutout} with varying Cutout length. We present the error rates of the ResNet-18 model on the CIFAR-10 test set.
     * means the published results in~\cite{keepaugment}.}
     \label{table:comparison-Kc}
     \begin{tabular}{c|ccc}
     \toprule[1.5pt]
          Cutout Length& Cutout~\cite{cutout} &KeepCutout~\cite{keepaugment} & IPF-Cutout \\ \hline
         8 & 4.7*&4.9*&\textbf{4.33} \\
         12  & 4.6* & 4.5*&\textbf{3.85}\\
         16  & 4.4* &3.9*&\textbf{3.58}\\
         20 & 4.5*& 4.0*&\textbf{3.53}\\
         24 & 5.1*&4.4*&\textbf{3.88} \\
       \bottomrule[1.5pt]
     \end{tabular}
\end{table}
\begin{table}[]
     \centering
     \caption{Additional comparison with KeepAugment~\cite{keepaugment} on CIFAR-10 and CIFAR-100, including both KeepCutout and KeepAutoAugment (KeepAA). * means the published results in~\cite{keepaugment}.}
     \label{table:comparison-KA}
     \begin{tabular}{c|lll}
     \toprule[1.5pt]
          CIFAR-10~\cite{cifar-10} & ResNet-18  & ResNet-110 & WRN28-10  \\ \hline
           Cutout~\cite{cutout} & 4.40* &5.20*  & 3.10* \\
         KeepCutout~\cite{keepaugment} & 3.80* & 4.50* &2.70*  \\
         IPF-Cutout &\textbf{3.14}&\textbf{3.37}&\textbf{2.57}\\ \hline
        CIFAR-100~\cite{cifar}  & ResNet-18  & WRN28-10 & Shake-26-32 \\ \hline
         AutoAugment~\cite{autoaugment} &20.22 & 17.10 &17.71 \\
         KeepAA~\cite{keepaugment} &19.70&17.10& 17.80 \\
         IPF-AA &\textbf{19.62}&\textbf{16.24}&\textbf{17.38} \\
       \bottomrule[1.5pt]
     \end{tabular}
\end{table}
\subsection{Comparison with KeepAugment~\cite{keepaugment}}
We compare our proposed framework and KeepAugment.
Our primary focus is on their application in conjunction with Cutout~\cite{cutout} and AutoAugment~\cite{autoaugment} techniques, as KeepAugment primarily brings improvements upon these two methods.
To facilitate comparison, we rigorously follow the experimental settings suggested in~\cite{keepaugment}. 
As KeepAugment's source code for reproduction remains unavailable, we undertake its implementation based on an unofficial reproduction\footnote{https://github.com/cjf8899/KeepAugment\_Pytorch}, which yields competitive results compared to the published results in the original paper. 
For CIFAR-10, we directly use the reported results in~\cite{keepaugment} and follow its training settings for fairness. For CIFAR-100, which was not reported in the original paper, we reproduce the results under our experimental settings.

Table~\ref{table:comparison-Kc} presents the test accuracy with different $l$ values.
It can be seen that although KeepAugment has brought improvements to the performance of Cutout, our proposed framework consistently outperforms KeepAugment, yielding more significant improvements in the performance of Cutout.
Furthermore, Table~\ref{table:comparison-KA} presents the test accuracy on CIFAR-10 and CIFAR-100, employing various deep networks.
While both KeepAugment and IPF-RDA exhibit performance improvements concerning Cutout and AutoAugment, IPF-RDA manifests notably higher test accuracy, e.g., achieving 1.13\% accuracy improvement on CIFAR-10 using ResNet-110.

A plausible explanation for the robust performance of IPF-RDA resides in its preservation of the most sensitive augmentation-relevant information within images.
This information is the most critical for classification decisions, thereby enhancing the overall robustness of DA methods.

  \begin{table}[]
     \centering
     \caption{Comparison with PuzzleMix~\cite{puzzlemix} and Co-Mixup~\cite{co-mixup} on CIFAR-100 and Tiny-ImageNet across multiple architectures. We use the reported results from~\cite{puzzlemix} (*). Per-epoch training time (s) is measured on CIFAR-100 using PreActResNet-18 with a 1-NVIDIA-2080-TI-GPU server. }
     \label{table:comparison-puzzle-comixup}
     \resizebox{0.99\columnwidth}{!}{
     \begin{tabular}{c|c|ccc|c}
     \toprule[1.5pt]
        & Tiny-Imagenet  & \multicolumn{3}{c|}{\text {CIFAR-100}}&Per-epoch  \\ \cline{2-5}
        &PreActR18& PreActR18 & WRN16-8 & ResNeXt29-4-24&costs\\ \hline
        PuzzleMix& 36.52* &20.62*&19.24*&21.12*&72.61s \\
        Co-Mixup& 35.85* &19.87*&19.15*&19.78*&192.37s \\  
        Ours &\textbf{34.33}&\textbf{18.91}&\textbf{18.67}&\textbf{19.23}&\textbf{42.31}s \\
       \bottomrule[1.5pt]
     \end{tabular}}
\end{table}
\subsection{Comparison with Saliency-Guided DA}
We compare our proposed framework with PuzzleMix and Co-Mixup on CIFAR-100 and Tiny-ImageNet, closely following the experimental settings and architectures used in~\cite{puzzlemix,co-mixup}.
As shown in Table~\ref{table:comparison-puzzle-comixup}, our method consistently achieves better performance across datasets and architectures.
In addition to its performance advantage, our method offers better training efficiency due to its lightweight design, incurring minimal extra cost during online training. 
We also account for the offline cost of the CDIEA module, which requires only about 2.5 GPU hours (Table~\ref{table:CDIEA-training-cost}).
Notably, this offline cost is a one-time preprocessing step per dataset. 
As a result, when training PreActResNet-18 for 300 epochs, the total GPU time for our method is approximately 6.03 hours, which is comparable to PuzzleMix (6.05 hours) and markedly lower than Co-Mixup (16.03 hours).
By decoupling a lightweight offline stage from efficient online training, our framework achieves a better balance between practicality and scalability.

\subsection{Analytical Results}\label{analysis_res}

\subsubsection{Visualization of Augmented Data}\label{sec:vis_augmented_data}
 To facilitate a comprehensive comparison of the augmented samples by different methods, we offer a visual analysis as depicted in Fig.~\ref{vis-of-augmented-data}.
It can be observed that the original augmentation methods may impact the critical information crucial for the accurate classification of images, resulting in the introduction of noisy augmented samples during the augmentation process.
Conversely, IPF-RDA is capable of effectively preventing critical information from being removed or recovering the damaged critical information while concurrently upholding the diversity intrinsic to the augmented data.
Therefore, IPF-RDA can enhance the robustness of data augmentation methods and further unleash their full potential for improved performance and generalization in various applications.


\subsubsection{t-SNE Visualization of Deep Features}
To perform a comparative analysis between models employing and not employing IPF-RDA, we visualize the learned deep features derived from the CIFAR-10 dataset.
To this end, we leverage the t-SNE algorithm~\cite{tsne}, which provides a visual representation to compare the performance of different models. 
Since automatic augmentation methods typically obtain stronger feature extraction capabilities~\cite{randaugment,trivialaugment}, we utilize a representative method, Fast-AutoAugment (FAA), to train the ResNet-50 models.
The results of the R-50 models trained with FAA and IPF-FAA are presented in Fig.~\ref{fig:tsne}\subref{faa} and Fig.~\ref{fig:tsne}\subref{ipf-faa}, respectively.

It can be observed intuitively that, using IPF-RDA, the distribution of different clusters presents superior inter-cluster separation and intra-cluster compactness, effectively enhancing the discriminative capability of learned features.
In order to accurately quantify these clustering results, the Dunn index (DI) is utilized, which is a quantitative metric considering both inter-cluster and intra-cluster distance.
DI is defined as follows:
\begin{equation}
    D I=\frac{\min _{1 \leq i \neq j \leq m} \delta\left(C_i, C_j\right)}{\max _{1 \leq j \leq m} \Delta_j}
\end{equation}
where separation $\delta\left(C_i, C_j\right)$ is the inter-cluster distance metric between clusters $C_i$ and $C_j$, and compactness $\Delta_j$ calculates the mean distance between all pairs in each cluster.
Therefore, a higher DI means better clustering.
 As is shown in Fig.~\ref{fig:tsne} using IPF-RDA, the clustering results achieve much higher DI values.
Specifically, models trained with the original FAA exhibit a DI value of $2.04\times10^{-6}$, while the DI value with IPF-FAA escalates to $3.28\times 10^{-5}$.
Therefore, with the integration of our framework, the DI value increases rapidly, indicating enhanced clustering performance.
Meanwhile, the enhanced clustering performance demonstrates a notable enhancement in the performance of FAA when integrated into our framework.

 \begin{table}[]
     \centering
     \caption{Transferred test accuracy on CIFAR-10. IPF-* means the corresponding methods are integrated into our framework. }
     \label{table:transfer-learning}
     \begin{tabular}{l|cc}
     \toprule[1.5pt]
          & Original & IPF-* \\ \hline
          Cutout~\cite{cutout} &92.41  & \textbf{92.80} \\
          RandomErasing~\cite{random_erasing} & 92.58 & \textbf{92.94} \\
          CutMix~\cite{cutmix} &92.78& \textbf{92.99} \\
          GridMask~\cite{gridmask} & 91.50 & \textbf{92.11} \\
          AdvMask~\cite{advmask} &91.51 & \textbf{91.73} \\
         AutoAugment~\cite{autoaugment} & 92.80 & \textbf{93.19} \\
         RandAugment~\cite{randaugment} & 92.69 & \textbf{92.84} \\
         FAA~\cite{fast_autoaugment} & 92.26 & \textbf{93.45} \\
         TrivialAugment~\cite{trivialaugment} & 92.75 & \textbf{93.39} \\
        \bottomrule[1.5pt]
     \end{tabular}
 \end{table}
\subsubsection{Transfer Learning}
In the data augmentation context, DA methods' effectiveness is also assessed with transfer learning~\cite{transfer-learning-1, transfer-learning-2}.
Therefore, we pre-train models on the CIFAR-100 dataset using various augmentations with and without IPF-RDA, followed by fine-tuning them on the CIFAR-10 dataset.
Theoretically, better augmentation methods enhance model performance and transferability to a greater extent.
Table~\ref{table:transfer-learning} illustrates the transferred test accuracy of various data augmentation methods with and without the utilization of IPF-RDA.
It can be observed that the incorporation of IPF-RDA yields consistent improvements in the transferred test accuracy.
Notably, the integration of IPF-RDA showcases its potential in improving the performance of regional deletion-based data augmentation methods by approximately 0.4\%, image mixing-based methods by 0.2\%,  and image-level methods by around 0.5\%.
Especially for Fast-AutoAugment, the transferred test accuracy experiences an impressive boost of 1.19\% with the integration of IPF-RDA.
Therefore, our framework significantly enhances the performance of the original data augmentation methods.

Additionally, the ResNet-50 models trained on the CIFAR-100 and transferred to CIFAR-10 using FAA and IPF-FAA are also presented in Fig.~\ref{fig:tsne}\subref{trans-faa} and Fig.~\ref{fig:tsne}\subref{trans-ipf-faa}, respectively.
In comparison between Fig.~\ref{fig:tsne}\subref{trans-faa} and Fig.~\ref{fig:tsne}\subref{trans-ipf-faa}, the DI value of original FAA is $4.43\times10^{-5}$, while that of IPF-FAA is as high as $7.58\times10^{-5}$. 
A higher DI value proves that our framework is highly effective in enhancing the transfer performance of data augmentation methods, resulting in models with stronger feature extraction capabilities.
\subsubsection{Values of The Proportion $Pr_{\Delta_g}$}\label{sec:prop}
\begin{figure}[]
	\centering
	\subfloat[ Values of $Pr_{\Delta_g}$
 ]{
		\includegraphics[width=0.48\columnwidth]{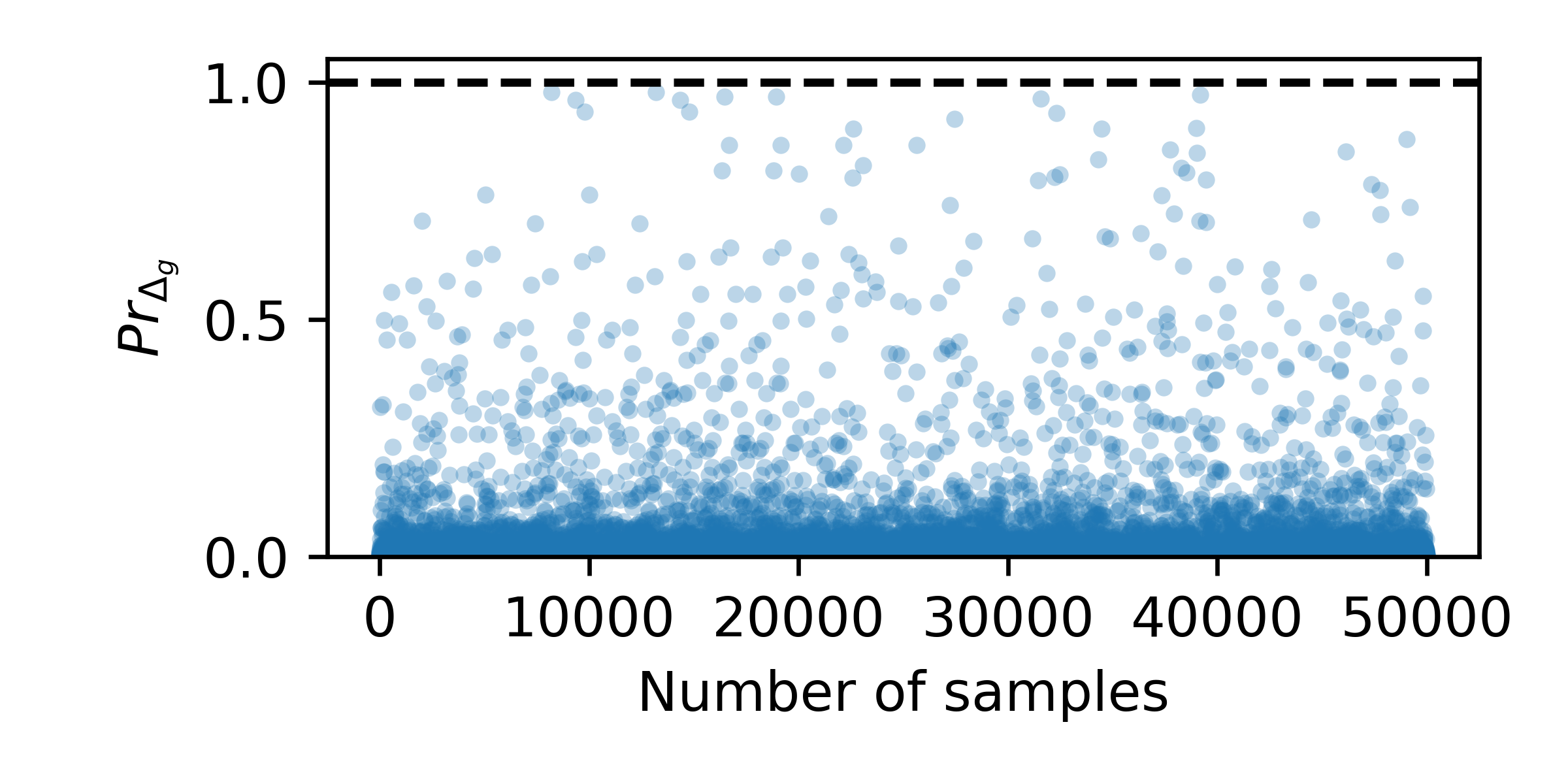}
        \label{fig:threotical-1}
	} 
	\subfloat[ CPD of $Pr_{\Delta_g}$
    ]{
		\includegraphics[width=0.48\columnwidth]{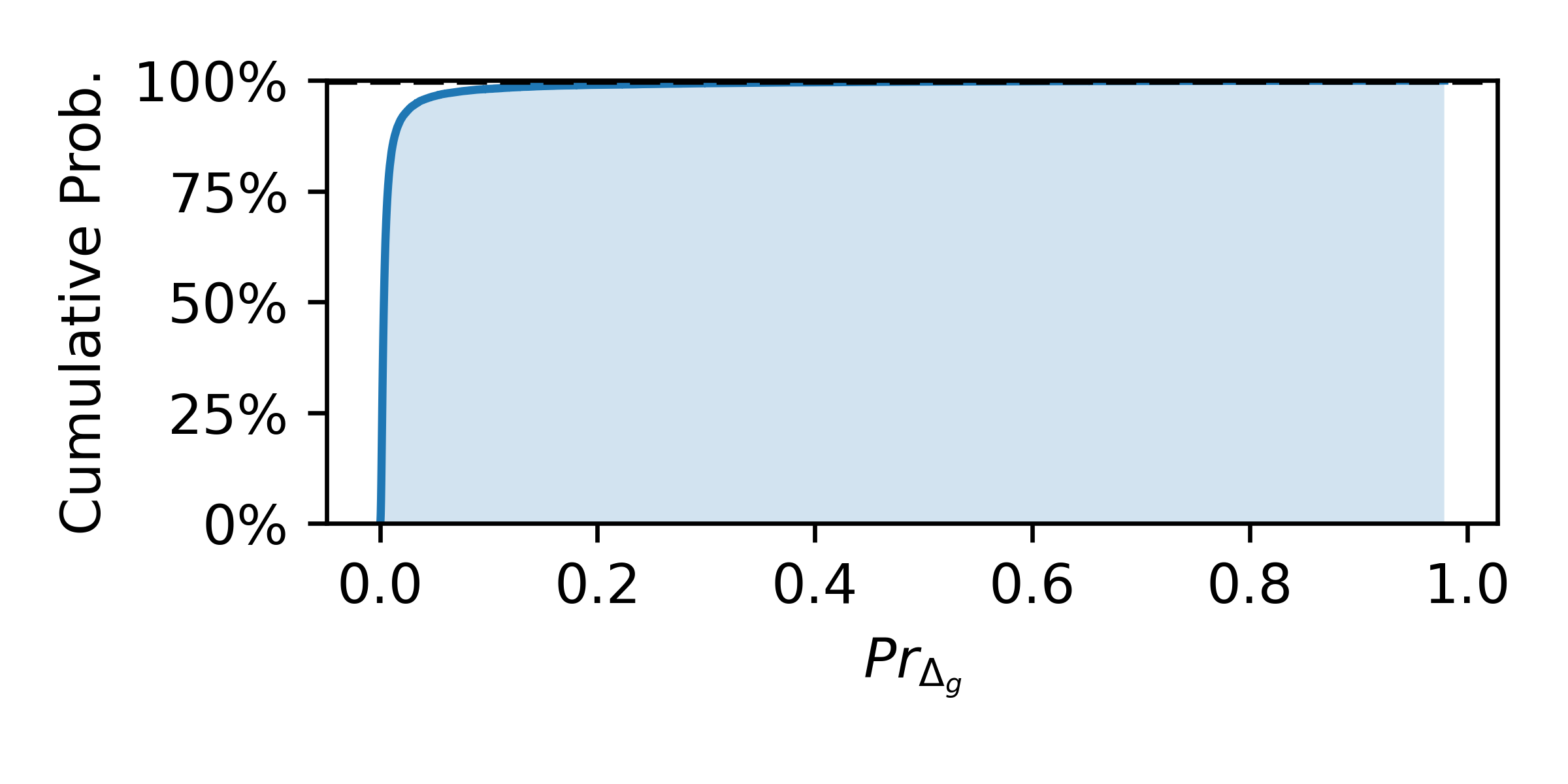}
        \label{fig:threotical-2}
	} 
	\caption{The proportion of the gradient change $\Delta_g$ w.r.t. $g_x$.}
	\label{fig:threotical}
\end{figure}

As aforementioned, the extent of alteration in the saliency map of the adversarial sample is bound by an upper limit, as established by Eq.~\eqref{eq:upperbound}.
This upper limit's relative smallness compared to the original gradient signifies a critical insight - the conventional saliency map lacks sensitivity to classification decisions and fails to encapsulate the most discriminative information inherent in images.
To assess the practical significance of this upper limit in relation to the original gradient, we undertake an empirical assessment by calculating the ratio of $\Delta_g$ to $g_x$, denoted as $Pr_{\Delta_g}= \left\| \Delta_g \right\| / \left\| g_x \right\|$.
Fig.~\ref{fig:threotical}\subref{fig:threotical-1} and Fig.~\ref{fig:threotical}\subref{fig:threotical-2} provide the visualizations of the $Pr_{\Delta_g}$ values and the cumulative probability distribution (CPD) of $Pr_{\Delta_g}$ utilizing the CIFAR-10 training set, respectively.
Evidently, it can be observed that the majority of samples exhibit $Pr_{\Delta_g}$ values below 1\% of the original gradient, with more than 98\% of instances falling within this threshold.
The gradient changes for the vast majority of samples are very small magnitudes, which are difficult to identify and capture.
These findings demonstrate that despite the significant changes in image classification results, the gradient changes observed in saliency maps are still notably small in magnitude.
Consequently, the saliency methods cannot be used as a basis for classification decisions, which also confirms the validity of our proposed approach.
 \begin{table}[]
     \centering
     \caption{Efficacy of CDIEA on various benchmark datasets with $\epsilon=8/255$. }
     \label{table:ASR}
     \begin{tabular}{l|c|c|c}
     \toprule[1.2pt]
     Dataset  & Success Rate(\%) & $\ell_0(\boldsymbol{\delta})$ & \% of pixels \\ \hline
      CIFAR-10~\cite{cifar} & 100 & 164.1 & 16.0\\
      CIFAR-100~\cite{cifar} & 100 & 172.5 & 16.8 \\ 
      Tiny-ImageNet~\cite{tiny} & 98.9 & 248.75 & 6.1 \\
      Market1501~\cite{market1501} & 100 & 1731.04 & 3.5 \\
      CUHK03~\cite{cuhk03} & 99.9 & 1889.9 & 3.8 \\
     \bottomrule[1.2pt]
     \end{tabular}
 \end{table}
\subsubsection{Class Activation Maps on Oxford Flower Dataset}
\begin{figure}[]
	\centering
	\begin{flushleft}
		\footnotesize ~~\quad Input \qquad~ Cutout \quad IPF-Cutout  \quad ~Input\qquad ~~~ RE  \qquad~~IPF-RE 
	\end{flushleft}
	\subfloat{
		\begin{minipage}{\Len\linewidth}
			\centering
			\setlength{\abovecaptionskip}{-0.01cm} 
			\includegraphics[width=\len \linewidth]{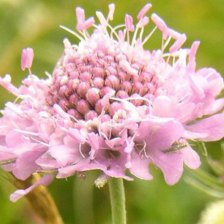}
		\end{minipage}
		\begin{minipage}{\Len\linewidth}
			\centering
			\setlength{\abovecaptionskip}{-0.01cm} 
			\includegraphics[width=\len \linewidth]{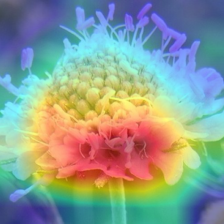}
		\end{minipage}
		\begin{minipage}{\Len\linewidth}
			\centering
			\setlength{\abovecaptionskip}{-0.01cm} 
			\includegraphics[width=\len \linewidth]{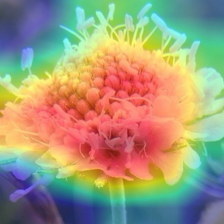}
		\end{minipage}
		\begin{minipage}{\Len\linewidth}
			\centering
			\setlength{\abovecaptionskip}{-0.01cm} 
			\includegraphics[width=\len \linewidth]{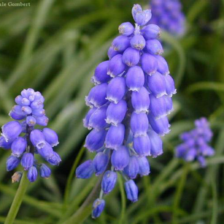}
		\end{minipage}
		\begin{minipage}{\Len\linewidth}
			\centering
			\setlength{\abovecaptionskip}{-0.01cm} 
			\includegraphics[width=\len \linewidth]{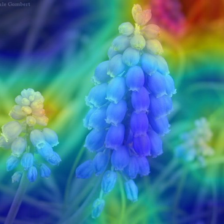}
		\end{minipage}
		\begin{minipage}{\Len\linewidth}
			\centering
			\setlength{\abovecaptionskip}{-0.01cm} 
			\includegraphics[width=\len \linewidth]{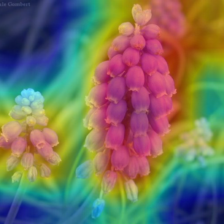}
		\end{minipage}
	}
 \vspace{\vlen mm }
	\newline
	\subfloat{
		\begin{minipage}{\Len \linewidth}
			\centering
			\includegraphics[width=\len \linewidth]{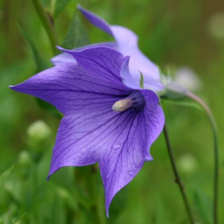}
		\end{minipage}
		
		\begin{minipage}{\Len\linewidth}
			\centering
			\includegraphics[width=\len \linewidth]{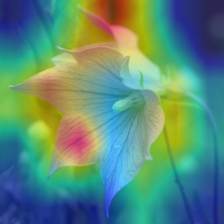}
		\end{minipage}
		\begin{minipage}{\Len\linewidth}
			\centering
			\includegraphics[width=\len\linewidth]{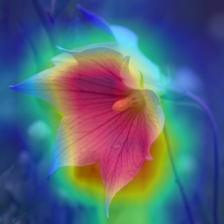}
		\end{minipage}
		\begin{minipage}{\Len \linewidth}
			\centering
			\includegraphics[width=\len\linewidth]{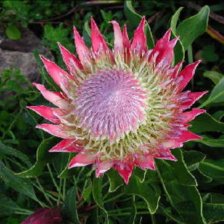}
		\end{minipage}
		\begin{minipage}{\Len\linewidth}
			\centering
			\includegraphics[width=\len\linewidth]{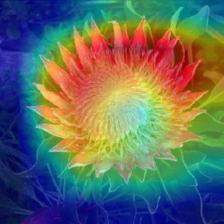}
		\end{minipage}
		\begin{minipage}{\Len \linewidth}
			\centering
			\includegraphics[width=\len\linewidth]{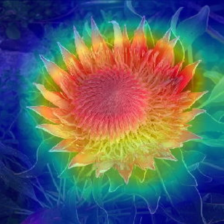}
		\end{minipage}
	}	
 \caption{Class activation mapping (CAM) for Oxford Flower Dataset~\cite{flower}.}
 \label{CAM}
\end{figure}

Since class activation maps (CAM) visualization can provide visual explanations for understanding model performance, we incorporate CAM visualization as a tool to analyze the discriminative regions used by models.
Since regional deletion-based methods can enhance the robustness of occlusion and increase the receptive field, we compare the performance before and after integrating these methods into our framework.
Fig.~\ref{CAM} illustrates the CAM generated by ResNet-50 trained with commonly used Cutout and Random Erasing on the Oxford Flower dataset~\cite{flower}.
After being integrated into our framework, these methods are more inclined to locate and highlight the most relevant parts of the main objects, while the background is almost ignored.
For instance, the original Cutout and RandomErasing's regions of interest cover only part of the main object (i.e., flower) and contain some irrelevant background information, indicating an overfitting problem.
In contrast, utilizing our framework, these methods' regions of interest cover almost the entire flower while ignoring much of the background, indicating better generalization ability.
To summarize, these visualizations underscore how our framework surpasses the efficacy of the original regional deletion-based data augmentation methodologies.

\subsubsection{Efficacy of CDIEA}
The results of CDIEA are crucial for balancing the diversity of augmented data and identifying the critical information.
To evaluate the efficacy of CDIEA in identifying the most class-discriminative information in images, we present the success rate of $\boldsymbol{x}+\boldsymbol{\delta}$ in influencing classification decisions, i.e., attack success rate (ASR).
Table~\ref{table:ASR} presents results on several benchmark datasets.
It can be observed that CDIEA achieves a nearly 100\% success rate on these datasets while maintaining a low $\ell_0$ norm, which means that only a marginal subset of candidate pixels are pinpointed as critical pixels.
As aforementioned, this preservation of limited but highly informative pixels significantly contributes to the diversity of augmented images, because only a small portion of the data will be considered as key information to preserve.
Consequently, by adaptively identifying and preserving the most class-discriminative regions within images, CDIEA ensures that augmented data not only retains its diversity but also provides valuable information for model training.

 \subsubsection{Efficacy of Rectangular Regions Preservation in IPF-RDA}
 In IPF-RDA, following~\cite{advmask,gridmask,cutout,trivialaugment,keepaugment}, we preserve a rectangular region containing the class-discriminative information.
 This design helps maintain the structured semantics in images while incurring negligible computational overhead.
 
To assess the effectiveness of this choice, we compare IPF-RDA with a point-level preservation variant based on image-level transformations (IPF-TA), in which the most class-discriminative points are retained after augmentation.
As shown in Table~\ref{table:rectangle-performance}, rectangular region preservation consistently outperforms the point-level preservation across different datasets and deep models by a large margin. 
These results highlight that maintaining contiguous rectangular areas better preserves the structural and semantic characteristics of images.
In contrast, point-level preservation may suffer from instability, as isolated points can be overly sensitive to adversarial attack pixels, thereby introducing misleading signals during training.
Moreover, we further discuss potential extensions to more advanced preservation strategies, such as superpixel-based regions, in Section~\ref{discussion}.
  \begin{table}[]
     \centering
     \caption{Efficacy of Rectangle-level preservation on CIFAR-10/100 using ResNet-18/50. We report error rate (\%).}
     \label{table:rectangle-performance}
     \begin{tabular}{l|cc|cc}
     \toprule[1.2pt]
    \multirow{2}{*}{Method}& \multicolumn{2}{c|}{CIFAR-10}& \multicolumn{2}{c}{CIFAR-100} \\ \cline{2-5}
      & ResNet-18 & ResNet-50& ResNet-18 & ResNet-50\\ \hline
     Point-level &5.60&5.20&24.10&21.32 \\
     Ours &\textbf{3.18} &\textbf{2.85}&\textbf{20.67} & \textbf{18.35} \\
     \bottomrule[1.2pt]
     \end{tabular}
 \end{table}
 
\subsubsection{Convergence Analysis of IPF-RDA}
 \begin{figure}[]
    \centering
    \includegraphics[width=0.75\columnwidth]{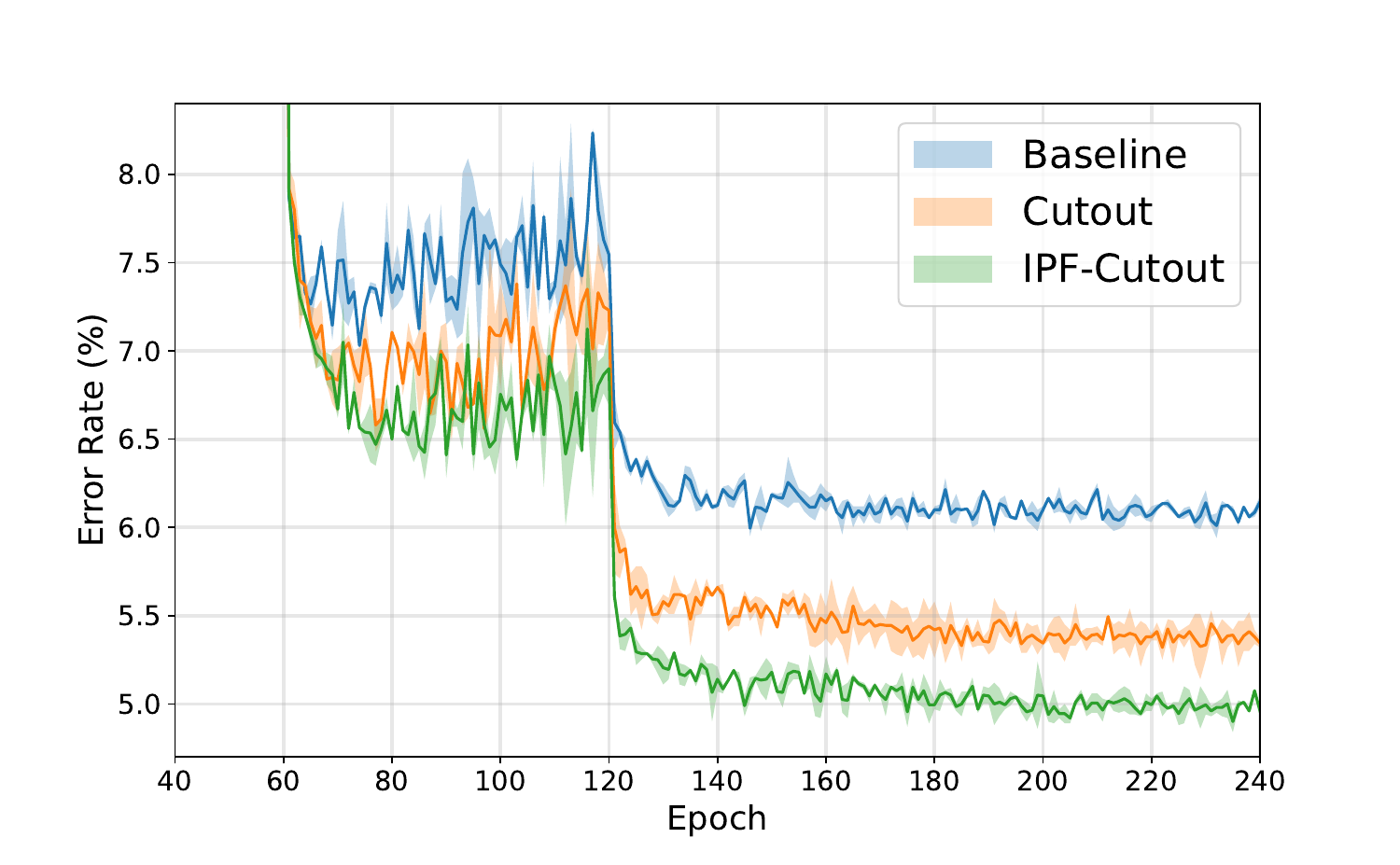}
    \caption{Curves of test errors on CIFAR-10 with ResNet-110.}
    \label{test-acc}
\end{figure}
While our method significantly contributes to model training, in this section, we demonstrate its ability to enhance the convergence rate of data augmentation methods, such as Cutout.
To more clearly present the dynamic evolution of test errors throughout the training process, following the experimental settings suggested in~\cite{isda}, we train ResNet-110 models on the CIFAR-10 dataset, utilizing a multi-step learning rate decay schedule.
The learning rate is initialized as 0.1 and multiplied by 0.2 at epochs 60, 120, 160, and 200. It is worth noting that all other experimental settings remain unchanged.

Fig.~\ref{test-acc} illustrates the test error curves corresponding to the baseline, Cutout, and IPF-Cutout, respectively. 
It can be observed that IPF-Cutout consistently showcases substantial improvements compared to both the baseline and the original Cutout technique.
This improved performance becomes particularly pronounced after the second learning rate drop, subsequently stabilizing the test errors of IPF-Cutout at notably lower levels.
By incorporating the Cutout augmentation within the proposed framework, we not only expedite the convergence rate but also enhance the overall performance beyond what can be achieved by the original Cutout approach.
This empirical evidence substantiates the remarkable efficacy of our framework, underscored by its capacity to facilitate more efficient training processes and substantially elevate overall performance.
 \begin{table}[t]
     \centering
     \caption{Comparison of training costs between CDIEA and popularly used deep models on CIFAR-100. Model costs are measured using a 2-NVIDIA-2080TI-GPU server, while CDIEA cost is measured on a single NVIDIA-2080TI GPU. We report total GPU hours (h).}
     \label{table:CDIEA-training-cost}
     \resizebox{0.99\columnwidth}{!}{
     \begin{tabular}{l|cccc|c}
     \toprule[1.2pt]
           & R-18 & R-50 & WRN-28-10 & Shake-2-32& CDIEA \\ \hline
        Training Cost (h) & 4.3 & 9.1 & 21.4 & 44.5 &2.5\\
        \bottomrule[1.2pt]
     \end{tabular}}
 \end{table}
 \subsection{Additional Offline Training Cost}\label{sec:offline-training-cost}
We analyze the additional offline training costs introduced by our CDIEA module.
For a better understanding, we compare it with the full training costs of widely-used deep models on CIFAR-100, including ResNet-18, ResNet-50, WRN-28-10, and Shake-Shake-2×32d, using a batch size of 256 and 8 parallel workers. 
As shown in Table~\ref{table:CDIEA-training-cost}, CDIEA requires only 2.5 GPU hours for training, significantly lower than the cost of training any of the standard models.  
More importantly, this is a one-time cost per dataset: once trained, CDIEA generates class-discriminative guidance that can be reused across different architectures and experiments without retraining.  
This design minimizes computational overhead and enhances the training efficiency of our method.

\begin{table}[t]
     \centering
     \renewcommand{\arraystretch}{1.2}
     \caption{Additional time consumption of IPF-RDA on CIFAR-10/100 datasets across different deep architectures. }
     \label{table:time-consumption}
     \resizebox{0.99\columnwidth}{!}{
     \begin{tabular}{c|c|c|c|c|c}
     \toprule[1.2pt]
   \multirow{3}{*}{\text { Networks }}&\multirow{3}{*}{\text { FLOPs}} & \multicolumn{4}{c}{\text { Additional Training Costs }} \\
   \cline {3-6}
   &&\multicolumn{2}{c|}{\text {Cutout~\cite{cutout}}} &\multicolumn{2}{c}{\text {AutoAugment~\cite{autoaugment}}} \\
   \cline {3-6}
    &&\multicolumn{1}{c|}{\text {CIFAR-10}}&\multicolumn{1}{c|}{\text {CIFAR-100}} & \multicolumn{1}{c|}{\text {CIFAR-10}} & \multicolumn{1}{c}{\text {CIFAR-100}}  \\ \hline
    ResNet-18\cite{resnet} & 1.82G  & +1.1\% & +1.4\% & +2.4\% & +3.1\% \\
    ResNet-50\cite{resnet} & 4.12G & +0.8\% & +1.4\% & +0.8\% & +0.1\% \\
    WRN-28-10\cite{wrn} & 5.25G & +1.0\% & +1.0\%& +1.0\% & +1.2\% \\
    Shake-26-2x32d\cite{2017Shake} & 426.89M & +0.1\% & +0.1\% & +9.6\% & +2.4\% \\
        \bottomrule[1.2pt]
     \end{tabular}}
\end{table}

\subsection{Additional Online Training Cost}\label{sec:training-cost}
Time costs are also a crucial metric for evaluating data augmentation methods.
In this section, we provide the additional training costs incurred by the IPF-RDA framework during the online training phase.
All models are trained on 2 NVIDIA RTX2080TI GPUs with a batch size of 256 and 8 parallel workers across five independent random trials.
Specifically, in Table~\ref{table:time-consumption}, we integrate the two most representative methods, Cutout and AutoAugment, into our framework on the CIFAR-10/100 datasets employing various deep networks.
It can be observed that IPF-RDA increases the computational costs of the original data augmentation methods by no more than 3\% across these most widely used deep networks on both CIFAR-10 and CIFAR-100 datasets.
Because CDIEA is decoupled from the online training procedure, it does not introduce additional training costs.
Therefore, the additional training costs are engendered by the random sampling of structural regions for information preservation.
Consequently, IPF-RDA does not significantly affect the training efficiency of these data augmentation approaches.
For a detailed discussion on the offline training costs of CDIEA, please refer to Section~\ref{discussion}.

\subsection{Ablation Study}~\label{sec:ablation-study}
\subsubsection{The Effect of CDIEA}
To further evaluate the effectiveness of CDIEA, we perform an ablation study in which CDIEA is replaced with alternative forms of key information: (1) corner points extracted using the Harris detector, and (2) saliency maps generated by a pretrained ResNet-50. 
The saliency-based information can be seamlessly incorporated into our framework.

As shown in Table~\ref{table:effect-of-CDIEA}, all variants of our information-preserving framework outperform the baseline, highlighting the overall benefits of the IPF in data augmentation.
Among them, CDIEA consistently achieves the lowest error rates across various architectures, validating its superior capability in identifying class-discriminative information.
These results demonstrate that, while pretrained saliency-based methods offer certain advantages, CDIEA provides more effective guidance when integrated into the IPF framework.
 \begin{table}[t]
     \centering
     \caption{The effect of CDIEA on CIFAR-10 using various architectures. We report error rates (\%).}
     \label{table:effect-of-CDIEA}
     \resizebox{0.85\columnwidth}{!}{
     \begin{tabular}{l|ccccc}
     \toprule[1.2pt]
          &  Resnet-18 & Resnet-50 & WRN-28-10  \\ \hline
        baseline &4.72 &4.34 &4.48 \\
        Corner Point &3.66 &3.44 & 2.59 \\
        Salient Point &3.52 &3.30 &2.55 \\ \hline
        Ours &\textbf{3.18} &\textbf{2.85} & \textbf{2.23} \\
        \bottomrule[1.2pt]
     \end{tabular}}
 \end{table}

\subsubsection{The Effect of Parameters}
\begin{figure}[t]
	\centering
	\subfloat[ Parameter $l$
 ]{
		\includegraphics[width=0.45\columnwidth]{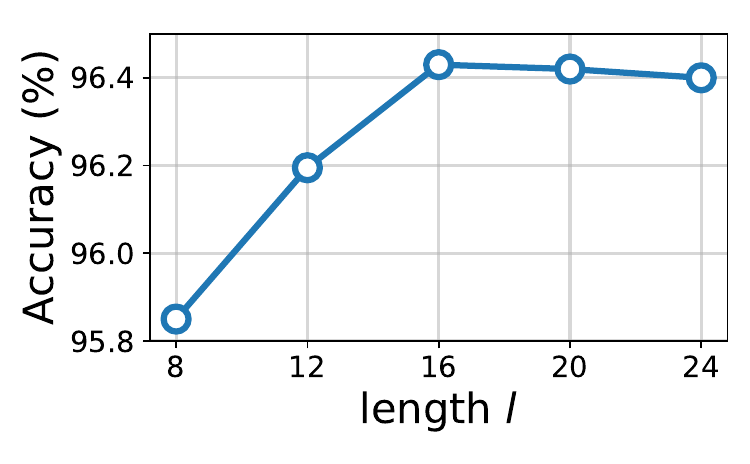}
        \label{ablation-length}
	} 
	\subfloat[ Parameter $\tau$
    ]{
		\includegraphics[width=0.45\columnwidth]{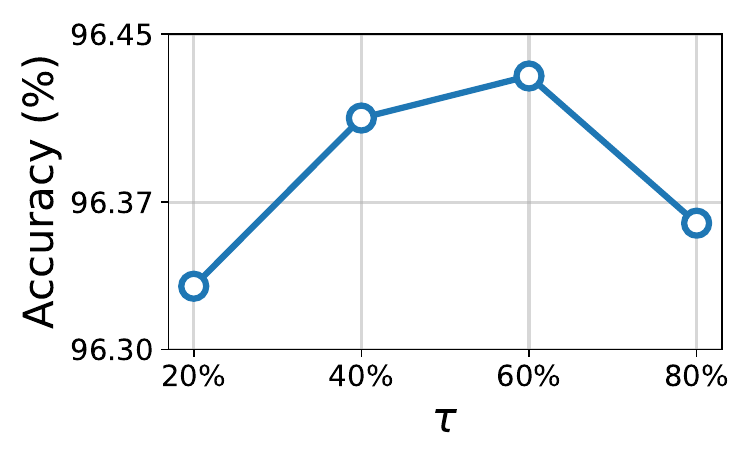}
        \label{ablation-tau}
	} 
	\caption{The ablation study of parameters for IPF-RDA.}
	\label{fig:ablation}
\end{figure}

To better understand the efficacy of two parameters in IPF-RDA, specifically denoted as $l$ and $\tau$, we embark on a series of ablation studies conducted on the CIFAR-10 dataset.
As demonstrated in Fig.~\ref{fig:ablation}\subref{ablation-length}, a controlled examination is carried out wherein $\tau$ is fixed as 60\%, while varying $l$ values spanning the range from 8 to 24.
It can be seen that the highest accuracy is obtained when $l=16$, which is consistent with the suggestions in~\cite{cutout}.
In Fig.~\ref{fig:ablation}\subref{ablation-tau}, we fix $l$ as 16 and employ various $\tau$ values from 20\% to 80\%.
It can be observed that IPF-RDA with $\tau=60\%$ performs best.
In fact, a larger value of $\tau$ indicates that the preserved regions contain information with higher importance scores. 
However, larger $\tau$ values lead to a reduced sample space size, limiting the diversity of augmented data and increasing the sampling consumption.
Meanwhile, a small value of $\tau$ leads to our method being insufficiently sensitive in capturing key information.
For a more comprehensive exploration encompassing diverse combinations of $\tau$ and $l$, please refer to Appendix~\ref{appendix-sec-ablation}.

\textbf{Parameter Settings}
The selection of the parameter $l$ is consistent with the suggestion in ~\cite{cutout}.
Expressly, we set $l$ to 8 for CIFAR-100 and 16 for others.
Moreover, adhering to the recommendations articulated in~\cite{keepaugment}, $\tau$ is set as a constant value of 60\% across all datasets.
Consequently, the parameters within the IPF-RDA are easy to configure, enhancing the practical applicability and ease of utilization. 

\section{Discussion and Future Work}\label{discussion}
In this work, we propose a novel information-preserving framework to preserve critical information for robust data augmentation, namely IPF-RDA.
The architecture of IPF-RDA consists of a two-step process.
It first identifies the most class-discriminative information, concurrently assigning corresponding importance scores.
Subsequently, this information forms the foundation for the seamless integration of various types of data augmentation methods, carefully categorized based on their operational characteristics.
Our experimental results demonstrate that IPF-RDA effectively enhances the robustness of data augmentation methods and substantially further improves the generalization performance of deep neural networks.
Thus, while data augmentation techniques have already proven their effectiveness in enhancing the performance of deep models, our framework strengthens their potential capabilities to a considerable extent.

While our extensive validation demonstrates that the proposed framework integrates well with a wide range of DA techniques, yielding consistent performance gains, incorporating image-level mixing methods, such as Mixup~\cite{mixup}, presents certain challenges. 
These methods generate convex combinations of image and labels, making it difficult to isolate and preserve class-discriminative information from the mixed content. 
Beyond image-level augmentation, feature augmentation methods~\cite{li2021simple,devries2017dataset} face similar issues, as pixel-level discriminative regions may not directly map to latent features. 
Extending our framework to address these challenges remains an interesting direction for future exploration.

Regarding spatial transformations-based DA, geometric deformation potentially misaligns the augmented images with the original class-discriminative features.
Our framework can be readily adapted to these cases by applying the same geometric transformation to the preserved regions, thereby maintaining spatial alignment. 
However, when the transformation is overly aggressive, the key information may be compromised.
For instance, when critical objects are displaced or their structural integrity is warped, the preserved regions may no longer contain meaningful semantics.
To mitigate such adverse effects, additional strategies, such as semantic-aware augmentation or dynamic region re-selection, may be necessary to ensure effective information preservation.

In addition, while CDIEA has proven effective in capturing critical information within images, it inevitably incurs additional computational costs. 
Similar to automatic DA methods that use reinforcement learning to pre-search the augmentation space, our framework is designed to strike an optimal balance between efficiency and performance.
By decoupling CDIEA from the online training process, we ensure that its impact on overall training time remains minimal. 
Once the one-time procedure is completed, no further CDIEA training or inference is needed. 
In the future, we envision a more lightweight variant that can be optionally integrated into the online training loop, offering an extra level of efficiency without altering the core benefits of the framework.

During augmentation, we leverage the rectangle-level regions for information preservation in IPF-RDA, which achieves a favorable balance between efficacy and efficiency~\cite{keepaugment,advmask,gridmask}.
While this design effectively maintains structured spatial information with minimal computational cost, more advanced alternatives, such as superpixel-based or semantic-segmentation-guided region selection, could align the preserved areas more closely with object boundaries and semantic parts, thereby enhancing the retention of fine-grained class-discriminative features based on our point-level importance scores. 
As such methods may inevitably introduce extra overhead, an interesting direction for future work is to explore lightweight, semantically aligned preservation strategies that maintain these benefits without compromising efficiency.

While our evaluation demonstrates that IPF-RDA consistently improves performance, its benefits may be compromised in scenarios with severe label noise. For example, we conduct experiments on CIFAR-100 with 50\% label noise. 
In this setting, the baseline, TrivialAugment, and IPF-TA achieve error rates of 36.35\%, 36.80\%, and 36.40\%, respectively.
These results indicate that although IPF-RDA can alleviate the performance drop, it cannot deliver substantial improvements in the presence of noisy data.
This is because preserving class-discriminative regions may inadvertently reinforce misleading features and learning signals tied to incorrect labels, thereby impairing training. 
Developing noise-robust preservation strategies, such as label uncertainty estimation, remains an interesting direction for future work.

 \section{Conclusion}
In this paper, we propose an information-preserving framework for robust data augmentation (IPF-RDA) with the aim of enhancing the robustness of data augmentation methods and further improving their performance.
The primary motivation behind IPF-RDA is the need to identify the class-discriminative pixels that are most vulnerable to perturbations; thus, these sensitive points are also very vulnerable to data augmentation operations.
Subsequently, our framework protects this sensitive information from being influenced after augmentations.
By preserving this critical information while not sacrificing the diversity of augmented data, our framework can boost the full potential of data augmentation approaches.
Experiment results on various computer vision benchmarks demonstrate the effectiveness and efficiency of the proposed method.
\bibliographystyle{IEEEtran}
\bibliography{tpami}

\begin{thebibliography}{10}
\providecommand{\url}[1]{#1}
\csname url@samestyle\endcsname
\providecommand{\newblock}{\relax}
\providecommand{\bibinfo}[2]{#2}
\providecommand{\BIBentrySTDinterwordspacing}{\spaceskip=0pt\relax}
\providecommand{\BIBentryALTinterwordstretchfactor}{4}
\providecommand{\BIBentryALTinterwordspacing}{\spaceskip=\fontdimen2\font plus
\BIBentryALTinterwordstretchfactor\fontdimen3\font minus
  \fontdimen4\font\relax}
\providecommand{\BIBforeignlanguage}[2]{{%
\expandafter\ifx\csname l@#1\endcsname\relax
\typeout{** WARNING: IEEEtran.bst: No hyphenation pattern has been}%
\typeout{** loaded for the language `#1'. Using the pattern for}%
\typeout{** the default language instead.}%
\else
\language=\csname l@#1\endcsname
\fi
#2}}
\providecommand{\BIBdecl}{\relax}
\BIBdecl

\bibitem{survey0}
M.~Xu, S.~Yoon, A.~Fuentes, and D.~S. Park, ``A comprehensive survey of image
  augmentation techniques for deep learning,'' \emph{Pattern Recognition}, p.
  109347, 2023.

\bibitem{survey1}
C.~Shorten and T.~M. Khoshgoftaar, ``A survey on image data augmentation for
  deep learning,'' \emph{Journal of big data}, vol.~6, no.~1, pp. 1--48, July
  2019.

\bibitem{survey2}
S.~Yang, W.~Xiao, M.~Zhang, S.~Guo, J.~Zhao, and F.~Shen, ``Image data
  augmentation for deep learning: A survey,'' \emph{arXiv preprint
  arXiv:2204.08610}, 2022.

\bibitem{cutout}
T.~DeVries and G.~W. Taylor, ``Improved regularization of convolutional neural
  networks with cutout,'' \emph{arXiv preprint arXiv:1708.04552}, 2017.

\bibitem{cutmix}
S.~Yun, D.~Han, S.~J. Oh, S.~Chun, J.~Choe, and Y.~Yoo, ``Cutmix:
  Regularization strategy to train strong classifiers with localizable
  features,'' in \emph{Proc. IEEE Int. Conf. Comput. Vis.}, 2019, pp.
  6023--6032.

\bibitem{autoaugment}
E.~D. Cubuk, B.~Zoph, D.~Mane, V.~Vasudevan, and Q.~V. Le, ``Autoaugment:
  Learning augmentation strategies from data,'' in \emph{Proc. IEEE Conf.
  Comput. Vis. Pattern Recognit.}, 2019, pp. 113--123.

\bibitem{randaugment}
E.~D. Cubuk, B.~Zoph, J.~Shlens, and Q.~Le, ``Randaugment: Practical automated
  data augmentation with a reduced search space,'' in \emph{Proc. Adv. Neural
  Inf. Process. Syst.}, vol.~33, 2020, pp. 18\,613--18\,624.

\bibitem{trivialaugment}
S.~G. M{\"u}ller and F.~Hutter, ``Trivialaugment: Tuning-free yet
  state-of-the-art data augmentation,'' in \emph{Proc. IEEE Int. Conf. Comput.
  Vis.}, 2021, pp. 774--782.

\bibitem{keepaugment}
C.~Gong, D.~Wang, M.~Li, V.~Chandra, and Q.~Liu, ``Keepaugment: A simple
  information-preserving data augmentation approach,'' in \emph{Proc. IEEE
  Conf. Comput. Vis. Pattern Recognit.}, 2021, pp. 1055--1064.

\bibitem{advmask}
S.~Yang, J.~Li, T.~Zhang, J.~Zhao, and F.~Shen, ``Advmask: A sparse adversarial
  attack-based data augmentation method for image classification,''
  \emph{Pattern Recognition}, vol. 144, p. 109847, 2023.

\bibitem{yang2025dynamic}
S.~Yang, P.~Ye, F.~Shen, and D.~Zhou, ``When dynamic data selection meets data
  augmentation,'' \emph{arXiv preprint arXiv:2505.03809}, 2025.

\bibitem{selectaugment}
S.~Lin, Z.~Zhang, X.~Li, and Z.~Chen, ``Selectaugment: Hierarchical
  deterministic sample selection for data augmentation,'' \emph{Proc. Assoc.
  Adv. Artif. Intell.}, vol.~37, no.~2, pp. 1604--1612, Jun. 2023.

\bibitem{yang2024clip}
S.~Yang, P.~Ye, W.~Ouyang, D.~Zhou, and F.~Shen, ``A clip-powered framework for
  robust and generalizable data selection,'' \emph{arXiv preprint
  arXiv:2410.11215}, 2024.

\bibitem{ma2022sage}
A.~Ma, N.~Dvornik, R.~Zhang, L.~Pishdad, K.~G. Derpanis, and A.~Fazly, ``Sage:
  Saliency-guided mixup with optimal rearrangements,'' \emph{arXiv preprint
  arXiv:2211.00113}, 2022.

\bibitem{saliency}
J.~Gu and V.~Tresp, ``Saliency methods for explaining adversarial attacks,''
  \emph{arXiv preprint arXiv:1908.08413}, 2019.

\bibitem{saliency_reliabiliey}
P.-J. Kindermans, S.~Hooker, J.~Adebayo, M.~Alber, K.~T. Sch{\"u}tt,
  S.~D{\"a}hne, D.~Erhan, and B.~Kim, ``The (un) reliability of saliency
  methods,'' \emph{Explainable AI: Interpreting, explaining and visualizing
  deep learning}, pp. 267--280, 2019.

\bibitem{SalientDN}
A.~Mahendran and A.~Vedaldi, ``Salient deconvolutional networks,'' in
  \emph{Proc. Eur. Conf. Comput. Vis.}, 2016.

\bibitem{sanity}
J.~Adebayo, J.~Gilmer, M.~Muelly, I.~Goodfellow, M.~Hardt, and B.~Kim, ``Sanity
  checks for saliency maps,'' \emph{Proc. Adv. Neural Inf. Process. Syst.},
  vol.~31, 2018.

\bibitem{gridmask}
P.~Chen, S.~Liu, H.~Zhao, and J.~Jia, ``Gridmask data augmentation,''
  \emph{arXiv preprint arXiv:2001.04086}, 2020.

\bibitem{has}
K.~K. Singh and Y.~J. Lee, ``Hide-and-seek: Forcing a network to be meticulous
  for weakly-supervised object and action localization,'' in \emph{Proc. IEEE
  Int. Conf. Comput. Vis.}, Oct 2017, pp. 3544--3553.

\bibitem{random_erasing}
Z.~Zhong, L.~Zheng, G.~Kang, S.~Li, and Y.~Yang, ``Random erasing data
  augmentation,'' in \emph{Proc. Assoc. Adv. Artif. Intell.}, vol.~34, no.~07,
  2020, pp. 13\,001--13\,008.

\bibitem{fast_autoaugment}
S.~Lim, I.~Kim, T.~Kim, C.~Kim, and S.~Kim, ``Fast autoaugment,'' in
  \emph{Proc. Adv. Neural Inf. Process. Syst.}, vol.~32.\hskip 1em plus 0.5em
  minus 0.4em\relax Curran Associates, Inc., 2019.

\bibitem{cifar}
A.~Krizhevsky, G.~Hinton \emph{et~al.}, ``Learning multiple layers of features
  from tiny images,'' 2009.

\bibitem{tiny}
P.~Chrabaszcz, I.~Loshchilov, and F.~Hutter, ``A downsampled variant of
  imagenet as an alternative to the cifar datasets,'' \emph{arXiv preprint
  arXiv:1707.08819}, 2017.

\bibitem{cuhk03}
W.~Li, R.~Zhao, T.~Xiao, and X.~Wang, ``Deepreid: Deep filter pairing neural
  network for person re-identification,'' in \emph{Proc. IEEE Conf. Comput.
  Vis. Pattern Recognit.}, 2014, pp. 152--159.

\bibitem{market1501}
L.~Zheng, L.~Shen, L.~Tian, S.~Wang, J.~Wang, and Q.~Tian, ``Scalable person
  re-identification: A benchmark,'' in \emph{Proc. IEEE/CVF Int. Conf. Comput.
  Vis.}, 2015, pp. 1116--1124.

\bibitem{yang2024investigating}
S.~Yang, S.~Guo, J.~Zhao, and F.~Shen, ``Investigating the effectiveness of
  data augmentation from similarity and diversity: An empirical study,''
  \emph{Pattern Recognition}, vol. 148, p. 110204, 2024.

\bibitem{yang2024entaugment}
S.~Yang, F.~Shen, and J.~Zhao, ``Entaugment: Entropy-driven adaptive data
  augmentation framework for image classification,'' in \emph{European
  Conference on Computer Vision}.\hskip 1em plus 0.5em minus 0.4em\relax
  Springer, 2024, pp. 197--214.

\bibitem{mixup}
H.~Zhang, M.~Cisse, Y.~N. Dauphin, and D.~Lopez-Paz, ``mixup: Beyond empirical
  risk minimization,'' in \emph{Proc. Int. Conf. Learn. Represent.}, 2018.

\bibitem{puzzlemix}
J.-H. Kim, W.~Choo, and H.~O. Song, ``Puzzle mix: Exploiting saliency and local
  statistics for optimal mixup,'' in \emph{International conference on machine
  learning}.\hskip 1em plus 0.5em minus 0.4em\relax PMLR, 2020, pp. 5275--5285.

\bibitem{co-mixup}
J.-H. Kim, W.~Choo, H.~Jeong, and H.~O. Song, ``Co-mixup: Saliency guided joint
  mixup with supermodular diversity,'' \emph{arXiv preprint arXiv:2102.03065},
  2021.

\bibitem{dgm}
B.~Fang, M.~Jiang, A.~Nagpure, and J.~Shen, ``Adaptive data augmentation with
  deep parallel generative models,'' in \emph{Proc. Int. Conf. Learn.
  Represent.}, 2020.

\bibitem{dgm2}
N.-T. Tran, V.-H. Tran, N.-B. Nguyen, T.-K. Nguyen, and N.-M. Cheung, ``On data
  augmentation for gan training,'' \emph{IEEE Transactions on Image
  Processing}, vol.~30, pp. 1882--1897, 2021.

\bibitem{isda}
Y.~Wang, G.~Huang, S.~Song, X.~Pan, Y.~Xia, and C.~Wu, ``Regularizing deep
  networks with semantic data augmentation,'' \emph{IEEE Trans. Pattern Anal.
  Mach. Intell.}, vol.~44, no.~7, pp. 3733--3748, 2021.

\bibitem{SM1}
A.~Mahendran and A.~Vedaldi, ``Salient deconvolutional networks,'' in
  \emph{Proc. Eur. Conf. Comput. Vis.}\hskip 1em plus 0.5em minus 0.4em\relax
  Springer, 2016, pp. 120--135.

\bibitem{SM2}
J.~Adebayo, J.~Gilmer, M.~Muelly, I.~Goodfellow, M.~Hardt, and B.~Kim, ``Sanity
  checks for saliency maps,'' \emph{Proc. Adv. Neural Inf. Process. Syst.},
  vol.~31, 2018.

\bibitem{SM3}
J.~Gu and V.~Tresp, ``Saliency methods for explaining adversarial attacks,''
  \emph{ArXiv}, vol. abs/1908.08413, 2019.

\bibitem{whySAA}
X.~Cheng, H.~Zhang, Y.~Xin, W.~Shen, and Q.~Zhang, ``Why adversarial training
  of re{LU} networks is difficult?'' in \emph{Proc. Int. Conf. Learn.
  Represent.}, 2023.

\bibitem{class_sense1}
C.~Szegedy, W.~Liu, Y.~Jia, P.~Sermanet, S.~Reed, D.~Anguelov, D.~Erhan,
  V.~Vanhoucke, and A.~Rabinovich, ``Going deeper with convolutions,'' in
  \emph{Proc. IEEE Conf. Comput. Vis. Pattern Recognit}, June 2015.

\bibitem{class_sense2}
M.~D. Zeiler and R.~Fergus, ``Visualizing and understanding convolutional
  networks,'' in \emph{Proc. Eur. Conf. Comput. Vis.}\hskip 1em plus 0.5em
  minus 0.4em\relax Springer, 2014, pp. 818--833.

\bibitem{autopruner}
J.-H. Luo and J.~Wu, ``Autopruner: An end-to-end trainable filter pruning
  method for efficient deep model inference,'' \emph{Pattern Recognition}, vol.
  107, p. 107461, 2020.

\bibitem{pgdl}
F.~Croce and M.~Hein, ``Sparse and imperceivable adversarial attacks,'' in
  \emph{Proc. IEEE Int. Conf. Comput. Vis.}, 2019, pp. 4724--4732.

\bibitem{fgsm}
Y.~Dong, F.~Liao, T.~Pang, H.~Su, J.~Zhu, X.~Hu, and J.~Li, ``Boosting
  adversarial attacks with momentum,'' in \emph{Proc. IEEE Conf. Comput. Vis.
  Pattern Recognit.}, 2018, pp. 9185--9193.

\bibitem{evaluation}
R.~Gontijo-Lopes, S.~Smullin, E.~D. Cubuk, and E.~Dyer, ``Tradeoffs in data
  augmentation: An empirical study,'' in \emph{Proc. Int. Conf. Learn.
  Represent.}, 2021.

\bibitem{cifar-10}
A.~Krizhevsky and G.~Hinton, ``Convolutional deep belief networks on
  cifar-10,'' \emph{Unpublished manuscript}, vol.~40, no.~7, pp. 1--9, 2010.

\bibitem{uda}
Q.~Xie, Z.~Dai, E.~Hovy, T.~Luong, and Q.~Le, ``Unsupervised data augmentation
  for consistency training,'' \emph{Proc. Adv. Neural Inf. Process. Syst.},
  vol.~33, pp. 6256--6268, 2020.

\bibitem{tsne}
L.~Van~der Maaten and G.~Hinton, ``Visualizing data using t-sne,'' \emph{J.
  Machine Learning Research}, vol.~9, no.~11, 2008.

\bibitem{mnist}
Y.~LeCun, C.~Cortes, and C.~J. Burges, ``The {MNIST} database of handwritten
  digits, 1998,'' \emph{URL http://yann. lecun. com/exdb/mnist}, vol.~10,
  no.~34, p.~14, 1998.

\bibitem{cam}
B.~Zhou, A.~Khosla, A.~Lapedriza, A.~Oliva, and A.~Torralba, ``Learning deep
  features for discriminative localization,'' in \emph{Proc. IEEE Conf. Comput.
  Vis. Pattern Recognit.}, 2016, pp. 2921--2929.

\bibitem{flower}
M.-E. Nilsback and A.~Zisserman, ``Automated flower classification over a large
  number of classes,'' in \emph{2008 Sixth Indian Conference on Computer
  Vision, Graphics \& Image Processing}, 2008, pp. 722--729.

\bibitem{resnet}
K.~He, X.~Zhang, S.~Ren, and J.~Sun, ``Deep residual learning for image
  recognition,'' in \emph{Proc. IEEE Conf. Comput. Vis. Pattern Recognit.},
  2016, pp. 770--778.

\bibitem{wrn}
S.~Zagoruyko and N.~Komodakis, ``Wide residual networks,'' in \emph{Proc. Brit.
  Mach. Vis. Conf.}\hskip 1em plus 0.5em minus 0.4em\relax BMVA Press,
  September 2016, pp. 87.1--87.12.

\bibitem{2017Shake}
X.~Gastaldi, ``Shake-shake regularization,'' \emph{arXiv preprint
  arXiv:1705.07485}, 2017.

\bibitem{shakedrop}
Y.~Yamada, M.~Iwamura, T.~Akiba, and K.~Kise, ``Shakedrop regularization for
  deep residual learning,'' \emph{IEEE Access}, vol.~7, pp. 186\,126--186\,136,
  2019.

\bibitem{densenet}
G.~Huang, Z.~Liu, L.~van~der Maaten, and K.~Q. Weinberger, ``Densely connected
  convolutional networks,'' in \emph{Proc. IEEE Conf. Comput. Vis. Pattern
  Recognit.}, July 2017.

\bibitem{resnext}
S.~Xie, R.~Girshick, P.~Doll{\'a}r, Z.~Tu, and K.~He, ``Aggregated residual
  transformations for deep neural networks,'' in \emph{Proc. IEEE Conf. Comput.
  Vis. Pattern Recognit.}, 2017, pp. 1492--1500.

\bibitem{FPB}
S.~Zhang, Z.~Yin, X.~Wu, K.~Wang, Q.~Zhou, and B.~Kang, ``Fpb: feature pyramid
  branch for person re-identification,'' \emph{arXiv preprint
  arXiv:2108.01901}, 2021.

\bibitem{ssl}
A.~Oliver, A.~Odena, C.~A. Raffel, E.~D. Cubuk, and I.~Goodfellow, ``Realistic
  evaluation of deep semi-supervised learning algorithms,'' in \emph{Proc. Adv.
  Neural Inf. Process. Syst.}, vol.~31.\hskip 1em plus 0.5em minus 0.4em\relax
  Curran Associates, Inc., 2018.

\bibitem{transfer-learning-1}
S.~J. Pan and Q.~Yang, ``A survey on transfer learning,'' \emph{IEEE Trans.
  Knowl. Data Eng.}, vol.~22, no.~10, pp. 1345--1359, May 2010.

\bibitem{transfer-learning-2}
F.~Zhuang, Z.~Qi, K.~Duan, D.~Xi, Y.~Zhu, H.~Zhu, H.~Xiong, and Q.~He, ``A
  comprehensive survey on transfer learning,'' \emph{Proc. IEEE}, vol. 109,
  no.~1, pp. 43--76, 2021.

\bibitem{li2021simple}
P.~Li, D.~Li, W.~Li, S.~Gong, Y.~Fu, and T.~M. Hospedales, ``A simple feature
  augmentation for domain generalization,'' in \emph{Proceedings of the
  IEEE/CVF international conference on computer vision}, 2021, pp. 8886--8895.

\bibitem{devries2017dataset}
T.~DeVries and G.~W. Taylor, ``Dataset augmentation in feature space,''
  \emph{arXiv preprint arXiv:1702.05538}, 2017.

\bibitem{sohn2020fixmatch}
K.~Sohn, D.~Berthelot, N.~Carlini, Z.~Zhang, H.~Zhang, C.~A. Raffel, E.~D.
  Cubuk, A.~Kurakin, and C.-L. Li, ``Fixmatch: Simplifying semi-supervised
  learning with consistency and confidence,'' \emph{Advances in neural
  information processing systems}, vol.~33, pp. 596--608, 2020.

\end{thebibliography}
\newpage
\clearpage
\appendix
\subsection{Proof of Lemma~\eqref{lemma1}}\label{appendix:lemma}
\begin{proof}
    We use the first-order Taylor expansion to decompose the gradient of the loss $w.r.t.$ the adversarial sample $\boldsymbol{x}+\boldsymbol{\delta}$ as:
\begin{equation}\label{eq:adv_gra}
    \nabla_{\boldsymbol{x}+\boldsymbol{\delta}}  = \nabla_{\boldsymbol{x}} + H_x \boldsymbol{\delta} + \mathcal{R}_2(\boldsymbol{\delta})
\end{equation}
where $H_x$ is the Hessian matrix defined as $H_x \stackrel{\text{def}}{=} \frac{\partial^2}{\partial x \partial x^T} \mathcal{L}(f(x), y)$, and $\mathcal{R}_2(\boldsymbol{\delta})$ is the terms of the perturbation $\boldsymbol{\delta}$ no less than the second order.
Based on Proposition~\ref{prop:whySAA} and substituting $\boldsymbol{\delta}$ in Eq.~\eqref{eq:adv_gra}, we have
\begin{equation}\label{eq:adv_gra-2}
    \begin{aligned}
        \nabla_{\boldsymbol{x}+\boldsymbol{\delta}_m} & = \nabla_{\boldsymbol{x}} + H_x (\sum_i \frac{\left(1+\beta \lambda_i\right)^m-1}{\lambda_i} \gamma_i v_i+\mathcal{R}_2(\beta)) \\ 
        &\quad + \mathcal{R}_2(\boldsymbol{\delta}_m) \\
       &\approx  \nabla_{\boldsymbol{x}} + H_x\sum_i \frac{\left(1+\beta \lambda_i\right)^m-1}{\lambda_i} \gamma_i v_i +H_x\mathcal{R}_2(\beta)  \\
    \end{aligned}
\end{equation}
Practically, in the context of the adversarial attack, to generate subtle perturbations, the $\ell_{\infty}$-norm of $\boldsymbol{\delta}$ is usually constrained below a much smaller constant; thus Eq.~\eqref{eq:adv_gra-2} holds.
\end{proof}
\subsection{Proof of Proposition~\eqref{pro1}}\label{appendix:prop1}
\begin{proof}
Combining Eq.~\eqref{eq:adv_gra} and Eq.~\eqref{eq:adv_gra-2} in Lemma~\ref{lemma1}, $\Delta_g$ can be calculated by 
\begin{equation}\label{eq:delta}
\begin{aligned}
    \Delta_g &=\nabla_{x+\boldsymbol{\delta}_m}-\nabla_{x} \\
            &=  H_x \boldsymbol{\delta} + \mathcal{R}_2(\boldsymbol{\delta}) \\
            &\approx \sum_i \frac{\left(1+\beta \lambda_i\right)^m-1}{\lambda_i}   \gamma_i H_x  v_i. 
\end{aligned}
\end{equation}
Since we have $H_x=V \Lambda V^T = \sum_i \lambda_i v_i v_i^T$, the term $H_x v_j (\forall j, 0 \le j \leq n)$ can be derived as
\begin{equation}\label{eq:H}
    \begin{aligned}
        H_x v_j &= \sum_i \lambda_i v_i v_i^T v_j\\
        &= \lambda_j v_j.
    \end{aligned}
\end{equation}
Substituting Eq.~\eqref{eq:H} back to Eq.~\eqref{eq:delta}, the change in gradient can be written as 
\begin{equation}\label{eq:delta_g}
    \begin{aligned}
        \Delta_g &= \sum_i  (\left(1+\beta \lambda_i\right)^m-1)\gamma_i v_i \\
        &= \sum_i  (\left(1+\beta \lambda_i\right)^m-1)   \nabla_{x}^Tv_i   v_i.
    \end{aligned}
\end{equation}
For a more idealized case, we consider the infinite-step attack with the infinitesimal step size,
\begin{equation}
    \begin{aligned}
      & \lim_{m \to +\infty} \sum_i  (\left(1+\beta \lambda_i\right)^m-1)   \nabla_{x}^Tv_i   v_i \\
       =& \lim_{m \to +\infty} \sum_i  ( (1+\frac{\beta m \lambda_i }{m})^m-1)   \nabla_{x}^Tv_i   v_i\\ 
       =&\lim_{m \to +\infty} \sum_i  ( (1+\frac{\alpha \lambda_i }{m})^m-1)   \nabla_{x}^Tv_i   v_i\\ 
       =&   \sum_i (e^{\alpha \lambda_k}-1)  \nabla_{x}^Tv_i   v_i \\
    \end{aligned}
\end{equation}
where $\alpha>0$ denotes the overall adversarial strength with the step number going to infinity.
Here, we have, 
\begin{align}
      ||\Delta_g|| &= ||\sum_i (e^{\alpha \lambda_i}-1)  \nabla_{x}^Tv_i   v_i  || \label{lab1} \\
      & \leq \sum_i \lvert (e^{\alpha \lambda_i}-1)\rvert \cdot \lvert \nabla_{x}^Tv_i \rvert \cdot ||v_i|| \label{lab2} \\
      & \leq \sum_i \lvert (e^{\epsilon \lambda_i}-1)\rvert \cdot \lvert \nabla_{x}^Tv_i \lvert \cdot ||v_i|| \label{lab3}
\end{align}
In the above, the Inequality~\eqref{lab2} follows from Minkowski's inequality.
Since the $\ell_\infty$-norm of attack is rigorously constrained by $\epsilon$ in Eq.~\eqref{problem}, $\alpha \leq \epsilon$ can be ensured.
The proof of Inequality~\eqref{lab3} is given in Appendix~\ref{appendix:proof1}.
\end{proof}

\subsection{Proof of Inequality~\eqref{lab3}}\label{appendix:proof1}
In this section, we prove Inequality~\eqref{lab3} in Section~\ref{analysis_res} of the main paper.

Inequality~\eqref{lab3} is as follows:
\begin{equation}
    \sum_i \lvert (e^{\alpha \lambda_i}-1) \rvert \cdot \lvert \gamma_i \rvert \cdot \left \| v_i \right \| \leq \sum_i \lvert (e^{\epsilon \lambda_i}-1)\rvert \cdot \lvert \gamma_i \rvert \cdot \left \| v_i \right \|
\end{equation}
where $\lambda_i, \gamma_i \in \mathbb{R}$, $\left \| v_i \right \| \geq 0$, and $0 \leq \alpha \leq \epsilon$.
To prove Inequality~\eqref{lab3}, it is equivalent to prove the following Inequality:
\begin{equation}
    \lvert e^{a x}-1\rvert \leq \lvert e^{b x}-1\rvert 
\end{equation}
where $x, y \in \mathbb{R}$ and $0\leq a\leq b$.
\begin{proof}
\begin{align}
    &\lvert e^{a x}-1\rvert \leq \lvert e^{b x}-1\rvert \\
   \Leftrightarrow \quad  &
   \begin{cases}
        e^{a x}-1 \leq e^{b x}-1 \quad \text{if } x \geq 0 \\
        e^{a x}-1 \geq e^{b x}-1 \quad \text{if } x \leq 0
    \end{cases}
   \end{align}
If $x\geq 0$, $e^{a x}\leq e^{b x} $ is trivial since $0\leq a\leq b$.
Similarly, it can be proven that $x \leq 0$ follows a similar line of reasoning.

Finally, Eq.~\eqref{lab3} is proven.
\end{proof}

\subsection{Reproducibility}
For reproduction, we will make available the source code, as well as the results of CDIEA on benchmark datasets.
We hope that this enables researchers to employ our information-preserving framework directly or to obtain class-discriminative information on personal datasets.

\subsection{Details of Datasets}
The CIFAR-10 dataset~\cite{cifar} consists of 60000 32$\times$32 colored natural images in 10 classes, with 6000 images per class.
There are 50,000 training images and 10,000 test images.
The CIFAR-100 dataset~\cite{cifar} consists of 100 classes containing 600 32$\times$32 colored natural images each.
There are 500 training images and 100 testing images per class. 
Tiny-ImageNet dataset~\cite{tiny} is a subset of the ILSVRC-2012 dataset.
It has 200 classes; each class has 500 training images and 50 validation images. 
All images are of size 64$\times$64 pixels.

\begin{table}[]
      \centering
     \caption{Error rates (\%) of different combinations of parameters $l$ and $\tau$ on CIFAR-10.}
     \label{appendix:ablation1}
     \begin{tabular}{c|cccc}
     \toprule[1.5pt]
         \diagbox{$l$}{$\tau$}&20\%&40\%&60\%&80\% \\ \hline 
         8 & 4.20&4.47&4.15&4.13\\
         12 &4.20&3.96&3.80&3.83\\
         16 &3.67&3.59&3.57&3.64\\
         20 &3.66&3.60&3.58&3.69\\
         24 &3.74&3.77&3.60&3.63\\
       \bottomrule[1.5pt]
     \end{tabular}
\end{table}
\begin{table}[]
      \centering
     \caption{Error rates (\%) of different combinations of parameters $l$ and $\tau$ on CIFAR-100.}
     \label{appendix:ablation2}
     \begin{tabular}{c|cccc}
     \toprule[1.5pt]
         \diagbox{$l$}{$\tau$}&20\%&40\%&60\%&80\% \\ \hline 
         8 & 20.76&20.58&20.53&20.58\\
         12 &21.29&20.55&20.68&21.02\\
         16 &22.01&21.04&20.85&21.07\\
         20 &22.51&21.42&21.70&21.65\\
         24 &22.74&22.67&22.40&22.81\\
       \bottomrule[1.5pt]
     \end{tabular}
\end{table}
\subsection{Additional Results of Ablation Studies}\label{appendix-sec-ablation}
Table~\ref{appendix:ablation1} and Table~\ref{appendix:ablation2} illustrate the test accuracy of IPF-Cutout with different combinations of parameters $l$ and $\tau$ on CIFAR-10 and CIFAR-100, respectively.
It can be observed that fixing $l$ values, $\tau=60\%$ achieves superior results, the same is true for $l=16$ on CIFAR-10.
On CIFAR-100, when $\tau=60\%$ and $l=8$, the test accuracy is competitive across different values of $\tau$ and $l$.

\subsection{Additional Ablation on More Augmentation Techniques}
\begin{figure}[t]
	\centering
	\subfloat[ Parameter $l$ - IPF-CutMix
 ]{
		\includegraphics[width=0.45\columnwidth]{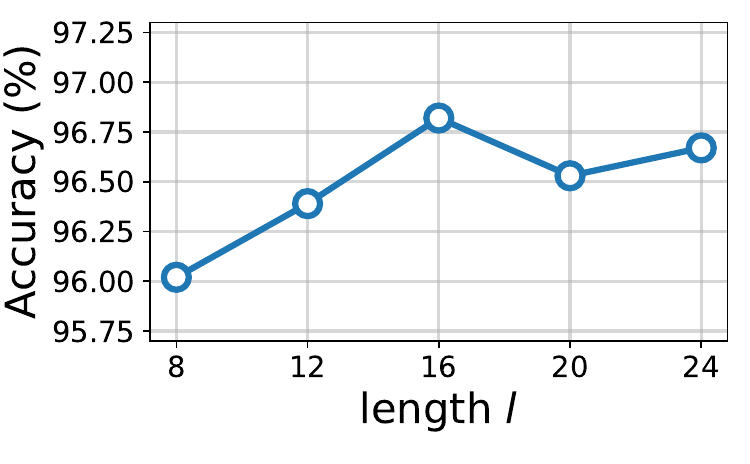}
        \label{ablation-length}
	} 
	\subfloat[ Parameter $\tau$ - IPF-CutMix
    ]{
		\includegraphics[width=0.45\columnwidth]{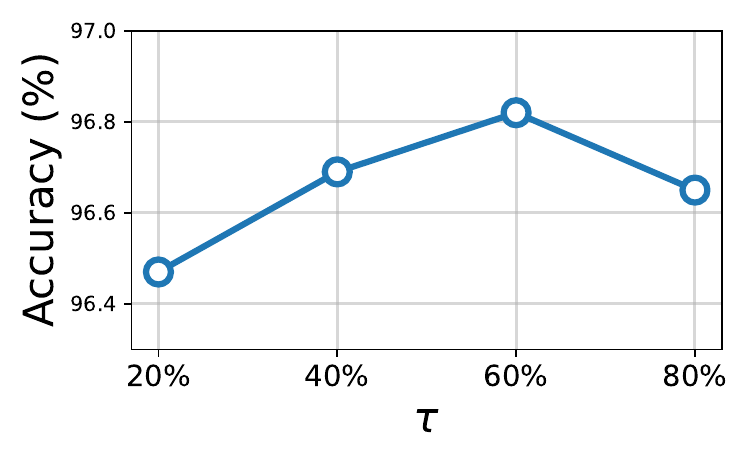}
        \label{ablation-tau}
	} 

    \subfloat[ Parameter $l$ - IPF-TA
 ]{
		\includegraphics[width=0.45\columnwidth]{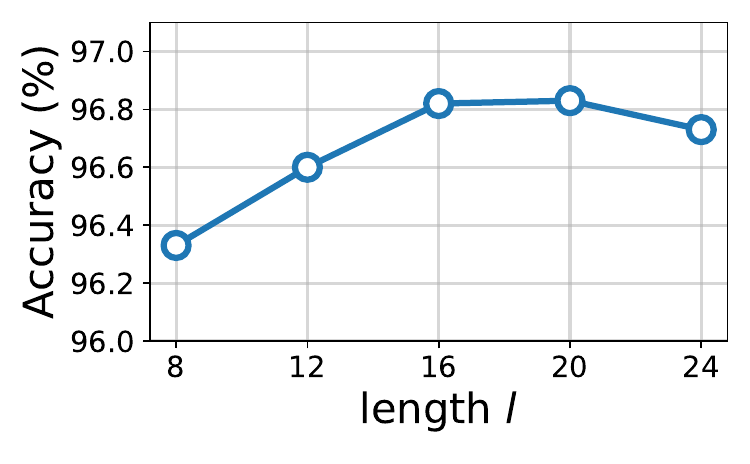}
        \label{ablation-length}
	} 
	\subfloat[ Parameter $\tau$ - IPF-TA
    ]{
		\includegraphics[width=0.45\columnwidth]{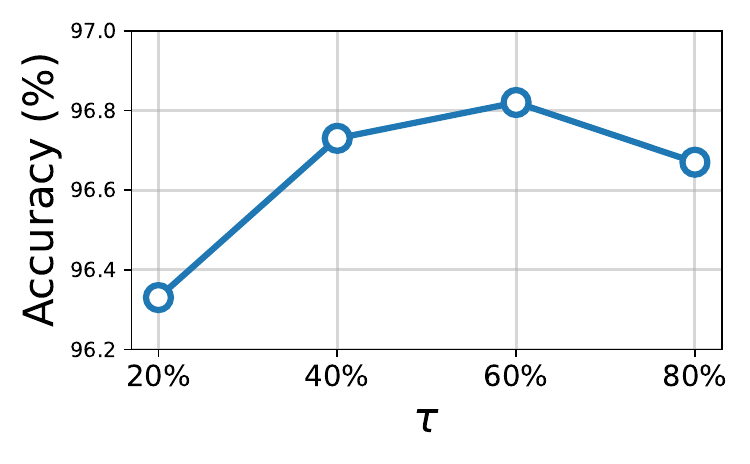}
        \label{ablation-tau}
	} 
    
	\caption{The ablation study for IPF-RDA across image mixing-based and image-level transformation-based methods on CIFAR-10. TA: TrivialAugment.}
	\label{fig:appendix-ablation-aug}
\end{figure}

In Section~\ref{sec:ablation-study}, we investigate the effect of $l$ and $\tau$ in IPF-Cutout and provide the parameter settings with minimal tuning overhead. 
Here, we further extend the parameter ablation to the other two augmentation categories, including both image-mixing-based and image-level transformation-based methods
For controlled examination, we fix $\tau$ as 60\% and vary $l$ values from 8 to 24.
Moreover, $l$ is fixed as 16 with varying $\tau$ values from 20\% to 80\%.
As shown in Figure~\ref{fig:appendix-ablation-aug}, parameter response trends are generally similar, and both parameters remain generally stable across different augmentation types.
This stability indicates that our method is robust to the choice of augmentation technique while incurring minimal parameter tuning cost.

\subsection{Extended Version of IPF-Cutmix}
In integrating CutMix into our framework in Section~\ref{sec:IPF}, our primary design goal is to retain the simplicity and efficiency of the original CutMix pipeline, introducing minimal modifications to ensure easy integration and low computational overhead.
Thus, in IPF-CutMix, we directly retain the class-discriminative information in $x_a$ and mix a random region from $x_b$ to maintain high data diversity.

In this section, we develop \emph{IPF-CutMix-v2}, an extended variant that preserves class-discriminative information from \textbf{both} $x_a$ and $x_b$. 
In this variant, the mixing process is designed to retain semantically informative content from both images by substituting a less critical region in $x_a$ with a highly critical region in $x_b$.
The label coefficients are computed based on the importance scores of the preserved region and the mixed region  (Eq.~\eqref{eq:lambda}).
As shown in Table~\ref{table:appendix-advanced-ipd-cutmix}, IPF-CutMix-v2 achieves comparable or slightly better performance than the original IPF-CutMix in most cases. 
This suggests that incorporating class-discriminative content from both images can be beneficial, while also demonstrating the flexibility and effectiveness of our information-preserving framework across different mixing strategies.

\begin{table}[t]
     \centering
     \renewcommand{\arraystretch}{1.2}
     \caption{Further analysis on the IPF-Cutmix using CIFAR-10/100 across various deep architectures. We report the error rates (\%). }
     \label{table:appendix-advanced-ipd-cutmix}
     \resizebox{0.99\columnwidth}{!}{
     \begin{tabular}{c|ccc|ccc}
     \toprule[1.5pt]
     \multirow{2}{*}{Method} & \multicolumn{3}{c|}{\text {CIFAR-10}} & \multicolumn{3}{c}{\text {CIFAR-100}} \\ \cline{2-7}
     & R-18 & R-50 & WRN-28-10& R-18 & R-50 & WRN-28-10 \\ \hline
     CutMix &3.36&3.19&3.07 &19.55&18.66&17.33 \\
     IPF-CutMix &3.18&2.84&2.45 &19.08&17.37&16.10 \\
     IPF-CutMix-v2 &3.20&2.69&2.40 &18.76&17.19&16.00 \\
    \bottomrule[1.5pt]
     \end{tabular}}
\end{table}

\subsection{Additional Results on SSL}
\begin{table*}[h]
     \centering
     \renewcommand{\arraystretch}{1.2}
     \caption{Semi-supervised learning error rates on CIFAR-10/100 via integrating our proposed method into FixMatch across different settings. FixMatch uses RandAugment~\cite{randaugment}. * means the published results in~\cite{sohn2020fixmatch}.}
     \label{table:appendix-fixmatch}
     \resizebox{0.75\textwidth}{!}{
     \begin{tabular}{c|lll|lll}
     \toprule[1.2pt]
     & \multicolumn{3}{c|}{\text {CIFAR-10}} & \multicolumn{3}{c}{\text {CIFAR-100}} \\ \cline{2-7}
    Method & 40 labels  & 250 labels & 4000 labels & 400 labels & 2500 labels & 10000 labels \\ \hline
    FixMatch &13.81&5.07&4.26 &48.85& 28.29&22.60 \\
    IPF-FixMatch &\textbf{13.10} \scriptsize{$\downarrow$ 0.71}&\textbf{4.78} \scriptsize{$\downarrow$ 0.29}&\textbf{4.09} \scriptsize{$\downarrow$ 0.17}&\textbf{45.17} \scriptsize{$\downarrow$ 3.68}&\textbf{26.35} \scriptsize{$\downarrow$ 1.94}&\textbf{21.46} \scriptsize{$\downarrow$ 1.44} \\
    
    \bottomrule[1.2pt]
     \end{tabular}}
\end{table*}
In Section~\ref{sec:ssl}, we demonstrate that IPF-RDA can be seamlessly integrated into data augmentation to enhance the performance of semi-supervised learning algorithms~\cite{uda,ssl}.
In this section, we extend the application of our proposed framework to FixMatch~\cite{sohn2020fixmatch}, another widely adopted SSL framework with strong performance.

Following standard SSL evaluation protocols and training settings~\cite{uda,sohn2020fixmatch}, we integrate IPF-RDA into FixMatch and conduct experiments on both CIFAR-10/100 across various settings. 
We report the average results across three independent trials.
As shown in Table~\ref{table:appendix-fixmatch}, while FixMatch can already achieve strong performance in SSL tasks, our framework can further consistently enhance its performance by a large margin across various SSL settings and datasets.
The results highlight the robustness and broad applicability of our method.


\vfill

\end{document}